\documentclass[10pt, a4paper, logo]{googledeepmind} %
\usepackage{times}

\usepackage{hyperref}
\usepackage[leftcaption]{sidecap}

\usepackage{hyperref}
\usepackage{url}
\usepackage{booktabs}
\usepackage{graphicx}
\usepackage{subcaption}
\usepackage{amssymb}
\usepackage[most]{tcolorbox}
\usepackage[table]{xcolor}
\tcbuselibrary{breakable} 
\usepackage{amsmath}
\usepackage{siunitx}
\usepackage{colortbl}

\usepackage{microtype}
\usepackage{color}
\usepackage{wrapfig}
\usepackage{natbib}
\usepackage{colortbl}
\usepackage{microtype}
\usepackage{graphicx}
\usepackage{booktabs} 
\usepackage{array}
\usepackage{textcomp}
\usepackage{stfloats}
\usepackage{float}
\usepackage{verbatim}
\usepackage[ruled, linesnumbered, lined]{algorithm2e}
\usepackage{multirow}
\usepackage{enumitem}

\definecolor{BestColor}{HTML}{C8E6C9}  
\definecolor{SecondBestColor}{HTML}{FFF9C4} 

\usepackage{tcolorbox}
\usepackage{amsmath,amsfonts}

\definecolor{ggg}{RGB}{26,179,0}
\definecolor{rrr}{RGB}{179,0,0}
\definecolor{oodc}{RGB}{31,73,121}
\definecolor{idc}{RGB}{68,142,68}

\definecolor{mygray}{gray}{0.9}

\def\Bias#1#2{\bm{b}}

\newtcolorbox{examplebox}[2][]{ 
    breakable, 
    enhanced, 
    colback=white, 
    colframe=cyan, 
    coltitle=white, 
    fonttitle=\bfseries, 
    title=#2, 
    overlay middle={\draw[cyan, line width=1pt](frame.south west)--(frame.south east);}, 
    overlay last={\draw[cyan, line width=1pt](frame.south west)--(frame.south east);}, 
    #1 
}

\usepackage[T1]{fontenc}
\usepackage{booktabs}      
\usepackage{graphicx}      
\usepackage[table]{xcolor} 
\usepackage{siunitx}       
\usepackage{etoolbox}      
\usepackage[normalem]{ulem}     

\definecolor{impcolor}{HTML}{2E8B57} 

\newcommand{\improvementstyle}[1]{$^{\textcolor{impcolor}{\tiny #1}}$}

\newcommand{\scoreimp}[2]{%
  \textbf{#1}%
  \ifstrequal{#2}{+0.0}{}{%
    \ifstrequal{#2}{0.0}{}{%
      \makebox[0pt][l]{\improvementstyle{#2}}%
    }%
  }%
}

\renewcommand{\titlefont}{\color{black}\normalfont\bfseries\fontsize{17}{20}\selectfont}
\title{Open Rubric System: Scaling Reinforcement Learning with Pairwise Adaptive Rubric}

\author[1,$*$]{Ruipeng Jia}
\author[1]{Yunyi Yang}
\author[1,4,$*$]{Wen Wang}
\author[2]{Yuxin Wu}
\author[1]{Yongbo Gai}
\author[3]{Siyuan Tao}
\author[1]{Mengyu Zhou}
\author[1]{Jianhe Lin}
\author[1]{Xiaoxi Jiang}
\author[1]{Guanjun Jiang}
\affil[1]{Qwen Large Model Application Team, Alibaba}
\affil[2]{Beijing University Of Posts and Telecommunications}
\affil[3]{Institute of Computing Technology, Chinese Academy of Sciences}
\affil[4]{University of Chinese Academy of Sciences}
\affil[$*$]{Equal contribution}

\begin{abstract}
  Scalar reward models compress multi-dimensional human preferences into a single opaque score, creating an information bottleneck that often leads to brittleness and reward hacking in open-ended alignment.
  We argue that robust alignment for non-verifiable tasks is fundamentally a \emph{principle generalization} problem: reward should be guided by explicit, auditable principles rather than compressed into a learned scalar function.
  To operationalize this view, we present the \textbf{Open Rubric System (OpenRS)}, a modular, rubrics-based \emph{LLM-as-a-Judge} framework built around \textbf{Pairwise Adaptive Meta-Rubrics (PAMR)} and lightweight \textit{Pointwise Verifiable Rubrics} (PVRs), which provide both hard-constraint guardrails and verifiable reward components when ground-truth or programmatic checks are available.
  OpenRS is governed by an explicit \emph{meta-rubric}---not a rubric itself, but a higher-level guideline that specifies \emph{how} rubrics should be generated, which criteria to prioritize, and which failure modes are non-negotiable---and instantiates concrete \emph{adaptive rubrics} on the fly by conditioning on the semantic differences between two candidate responses.
  It then performs criterion-wise \emph{pairwise} comparisons and aggregates criterion-level preferences externally, avoiding pointwise weighted scalarization while improving discriminability in open-ended settings.
  To keep principles consistent yet editable across various domains, we introduce a two-level meta-rubric refinement pipeline (automated evolutionary refinement for general principles and a reproducible human-in-the-loop procedure for domain principles), complemented with pointwise verifiable rubrics that act as both guardrails against degenerate behaviors and a source of verifiable reward for objective sub-tasks.
  Notably, the entire pipeline requires \emph{no judge-specific training}: the underlying judge model remains unmodified, and all evaluation policy lives in editable text.
  Finally, we instantiate OpenRS as reward supervision in pairwise RL training.
  Empirically, OpenRS achieves state-of-the-art results on four reward-modeling benchmarks (RM-Bench, JudgeBench, RewardBench v2, and PPE Preference),
  consistently outperforming strong open scalar reward model baselines and trained generative judges.
  In end-to-end policy optimization, replacing a scalar reward model with OpenRS yields stable gains on downstream RL evaluations.
  The open-source code is available at \url{https://github.com/Qwen-Applications/OpenRS}.
\end{abstract}

\begin{document}

\maketitle

\section{Introduction}
\label{sec:introduction}

Reinforcement learning from human feedback (RLHF) has become a standard paradigm for aligning large language models (LLMs) to follow instructions and exhibit helpful behaviors \citep{ouyang2022training}.
Typically, RLHF pipelines rely on a scalar reward model (RM) that compresses rich, multi-faceted human preferences into a single number.
While in practice, scalar RMs may learn shortcuts and become miscalibrated,
creating opportunities for reward hacking where policies exploit spurious cues rather than improving true response quality \citep{zhong2025comprehensive,mckee2024honesty}.

Recent progress on reinforcement learning with verifiable rewards (RLVR) demonstrates that when rewards can be grounded in objective checks (e.g., correctness or format constraints) for math or coding tasks,
reinforcement learning can scale capabilities without relying on a learned scalar judge \citep{guo2025deepseek}.
For general, open-ended tasks, reward specification is intrinsically difficult because the desired behavior spans multiple, often competing dimensions,
and there is typically no explicit reference answer or ground-truth signal.
One line of work follows the \emph{LLM-as-a-Judge} paradigm \citep{zheng2023judgingllm}, and attempts to train generative reward models or specialized judges from human or distilled preferences\citep{mahan2024generative,ye2024beyond},
and recent trained GenRM-style methods further incorporate chain-of-thought rationales and inference-time scaling to improve transparency and generalization \citep{liu2025inference,anugraha2025r3,yu2025rewardanything,whitehouse2025j1}.
Despite recent progress, these GenRM pipelines remain limited by their reliance on noisy, synthetic, and underspecified preference supervision,
making them costly to maintain and iterate across broader domains.
As optimization proceeds, the judge can exploit dataset-specific shortcuts and inadvertently erode the base model's general evaluation competence.
Additionally, the underlying evaluation principles are ultimately implicitly encoded rather than explicitly specified,
 which leads to scalar-RM-like brittleness under OOD shifts despite textual rationales.

A number of benchmarks leverage detailed rubrics generated by experts to provide a more structured and consistent evaluation in high-stakes professional domains \citep{arora2025healthbench,akyurek2025prbench,wang2025profbench}.
Building on this idea, an increasingly popular line of work attempts to automatically construct sophisticated rubrics as training-time rewards to extend RLVR-style optimization to more open-ended domains.
These rubric-based methods mostly rely on static rubrics that either synthesized by powerful LLMs \citep{gunjal2025rubricsrewards,huang2025reinforcement,ye2025selfrewarding,bi2025reward}, or induced from preference data \citep{liu2026openrubrics,li2026rubrichub},
which proven to be effective when the target response is well-scoped and can be anchored by expert standards (e.g., instruction following or high-stakes professional domains).
Recent work attempts to refine rubrics or eliciting criteria in a semi-static or online fashion \citep{zhang2025chasing,rezaei2025onlinerubric,li2026rubrichub,zhou2025breaking}.
However, in most cases the final training signal is still produced by pointwise rubric scoring followed by weighted scalar aggregation,
which reintroduces absolute score calibration across prompts, criteria, and judge calls \citep{gunjal2025rubricsrewards,huang2025reinforcement,ye2025selfrewarding}.
This pointwise scalarization creates an intrinsic ceiling on discriminability in open-ended settings and remains vulnerable to reward gaming,
which can limit improvement on non-verifiable tasks and contribute to collapse-like dynamics noted in open-ended RL \citep{jia2025writingzero,zhang2026arenarl}.
Moreover, many rubric construction pipelines implicitly describe the features of an ``ideal'' reference-style answer; while such anchoring can be reasonable for domains with strong norms,
it becomes harder to generalize to broader open-ended generation where the relevant trade-offs are context-dependent.
Preference-derived rubric pipelines \citep{liu2026openrubrics,xie2025auto} improve discriminability but can entangle criteria with annotator- or model-specific preference directions, washing out subtle pairwise trade-offs before RL optimization.

\begin{figure}[t]
    \centering
    \includegraphics[width=\textwidth]{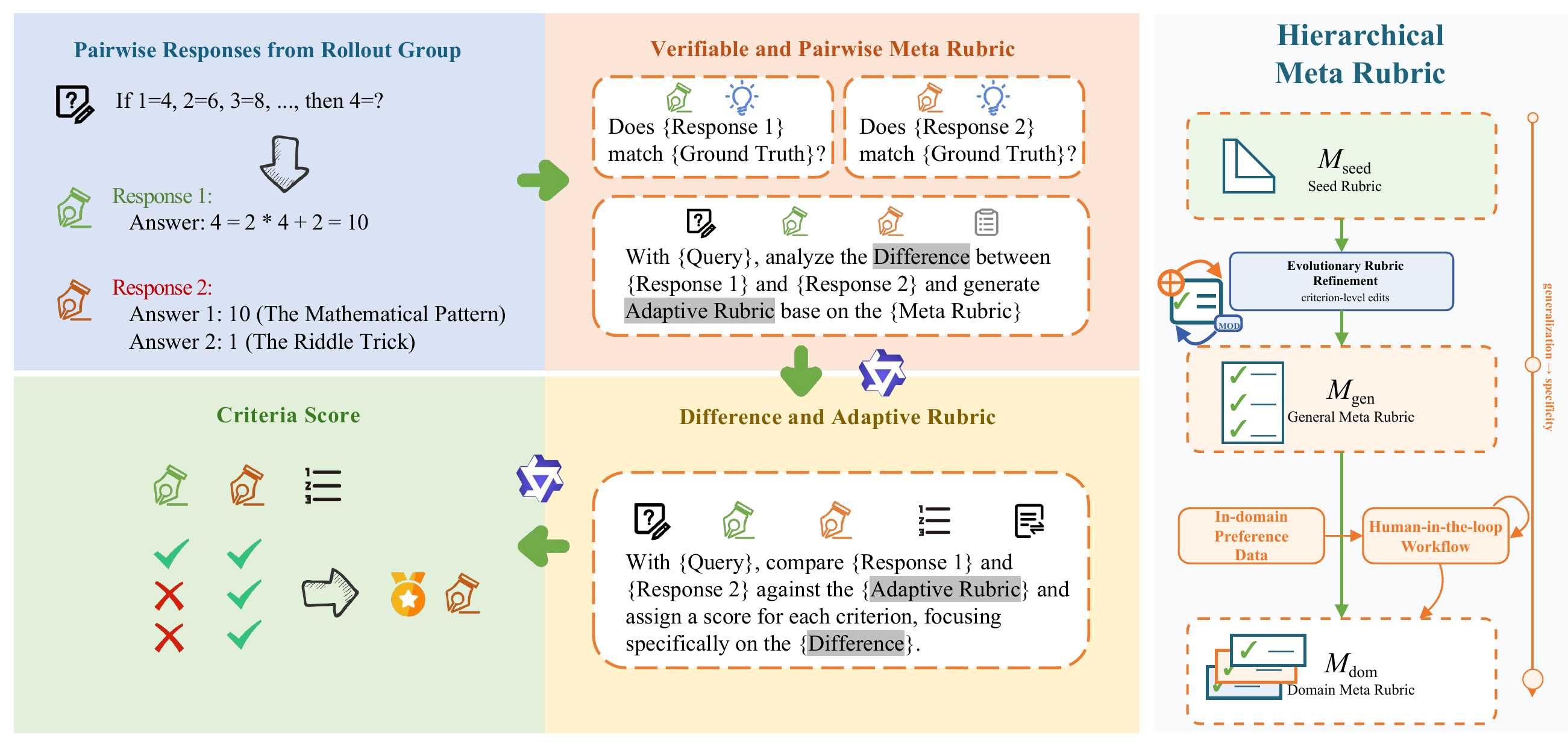}
    \captionsetup{justification=centering}
    \caption{Overview of OpenRS. Left: the OpenRS reward pipeline---given pairwise responses from the rollout group, OpenRS performs verifiable checking, difference-grounded adaptive rubric generation, and criterion-wise pairwise scoring. Right: hierarchical meta-rubric construction from seed rubric $\mathcal{M}_{\text{seed}}$ through evolutionary refinement to $\mathcal{M}_{\text{gen}}$, then domain adaptation via human-in-the-loop to $\mathcal{M}_{\text{dom}}$.}
    \label{fig:framework}
\end{figure}

In a nutshell, we still face the following challenges: (1) rewards from trained (scaler or generative) reward models are opaque, noisy, inefficient to iterate, and misalign with human value;
(2) the static specification and pointwise assessment of rubrics judge undermine the scalability of online RL in more general domains.
To address the above challenges, we argue that robust alignment in non-verifiable settings is fundamentally a \emph{principle generalization} problem:
reward should be guided by explicit, auditable principles rather than compressed into a learned scalar function, here we refer to such a principle specification as the \emph{meta-rubric}.
We propose \textbf{Pairwise Adaptive Meta-Rubrics (PAMR)}, which replaces trained scalar RMs as the source of the human-feedback signal with a general-purpose LLM judge---\emph{without any judge-specific training}---guided by an explicit \emph{meta-rubric}: not a rubric itself, but a higher-level guideline that specifies \emph{how} rubrics should be generated, which criteria to prioritize, and which failure modes are non-negotiable.
Specifically, rather than scoring a single response against an absolute rubric, PAMR conditions on the \emph{observable semantic differences} between two candidate responses to instantiate \emph{adaptive rubrics} on the fly,
performs rubric-wise \emph{pairwise} comparison, and aggregates criterion-level preferences externally for stable optimization.
Crucially, preference labels are used only for downstream meta-rubric refinement or validation, never for per-pair rubric induction, which keeps the instantiated criteria grounded in what actually differs between the two responses.
To keep the evaluation standard consistent with human principles while remaining explicit and editable, we refine the meta-rubric at two levels.
First, we refine the general meta-rubric via an automated evolutionary procedure.
Rather than learning a refinement policy for its own sake, we use optimization to discover a stronger, more transferable principle specification that generalizes across domains.
Second, we complement this with a domain meta-rubric refinement stage that supports human-in-the-loop iteration:
domain experts (or developers) can diagnose failure cases and update domain-specific principles in a reproducible, analytical manner without retraining the judge, enabling rapid iteration under distribution shifts.
Combined with lightweight pointwise verifiable rubrics that provide both verifiable reward components and hard-constraint guardrails (e.g., format constraints, unit tests, and ground-truth correctness checks) to prevent degenerate behaviors and reduce reward hacking during optimization,
we establish a modular rubric system, the Open Rubric System (OpenRS), that bridges RLVR and RLHF by combining pointwise verifiable rewards and constraints with meta-rubric-guided pairwise preference optimization for general domains.
Finally, we instantiate this design in a scalable RL framework that supports pairwise optimization.

Our contributions are as follows:
(1) We reframe robust reward supervision for non-verifiable tasks as a \emph{principle generalization} problem, and leverage an \emph{unmodified} LLM judge to instantiate \emph{adaptive rubrics} from explicit, optimized principles.
(2) We open-source a modular, rubrics-based LLM-as-a-judge system, OpenRS featuring Pairwise Adaptive Meta-Rubrics (PAMR) and lightweight pointwise verifiable rubrics, which outperforms prior scalar reward models and trained generative judges across four RM benchmarks.
(3) When used as reward signals and coupled with pairwise RL training, our method yields consistent gains on open-source policy benchmarks and in-house industrial evaluations.

\section{Preliminary}
\label{sec:preliminary}

\subsection{Group Relative Policy Optimization}

GRPO \citep{shao2024deepseekmath} is an efficient reinforcement learning algorithm that eliminates the need for a value function critic, as used in PPO \citep{schulman2017proximal}, by introducing group-relative normalization for advantage estimation.
Given a query $q$ from the dataset $\mathcal{D}$, the behavior policy $\pi_{\theta_{\text{old}}}$ samples a group of $G$ outputs $\{o_i\}_{i=1}^G$.
The advantage of the $i$-th response is subsequently derived by normalizing the rewards $\{R_i\}_{i=1}^G$ relative to the group statistics.
Similar to PPO, GRPO ensures training stability and sample efficiency by constraining policy updates $\pi_{\theta}$ within a trust region using a clipping mechanism.
Formally, GRPO optimizes the following objective:
\begin{equation}
  \begin{aligned}
    \mathcal{J}_{\text{GRPO}}(\theta) & = \mathbb{E}_{q \sim \mathcal{D}, \{o_i\}_{i=1}^G \sim \pi_{\theta_{\text{old}}}(\cdot|q)} \\
                                      & \left[\frac{1}{G} \sum_{i=1}^{G} \frac{1}{|o_i|} \sum_{t=1}^{|o_i|} \left( \min \left( r_{i,t}(\theta) \hat{A}_{i,t}, \ \text{clip} \left( r_{i,t}(\theta), 1-\varepsilon, 1+\varepsilon \right) \hat{A}_{i,t} \right) - \beta D_{\text{KL}}(\pi_{\theta} \| \pi_{\text{ref}}) \right) \right]
  \end{aligned}
\end{equation}

\begin{equation}
\label{grpo_advantage}
r_{i,t}(\theta) = \frac{\pi_{\theta}(o_{i,t} \mid q, o_{i,<t})}{\pi_{\theta_{\text{old}}}(o_{i,t} \mid q, o_{i,<t})} \qquad
\hat{A}_{i,t} = \frac{R_i - \text{mean}(\{R_i\}_{i=1}^G)}{\text{std}(\{R_i\}_{i=1}^G)}
\end{equation}
where $\hat{A}_{i,t}$ denotes the estimated advantage at time step $t$, $\varepsilon$ is the clipping hyperparameter for the importance sampling ratio $r_{i,t}(\theta)$, and $\pi_{\text{ref}}$ is the reference policy (typically initialized as $\pi_{\theta}$).

\subsection{Pairwise Evaluation and Bootstrapped Relative Policy Optimization}
\label{sec:pairwise_genrm_brpo}
\citet{jia2025writingzero} advocate for pairwise evaluation over pointwise scoring, particularly for non-verifiable tasks where absolute ground truth is elusive.
However, integrating pairwise feedback into group-based RL algorithms like GRPO presents a significant computational challenge.
To estimate the advantage of each response within a group of size $G$, a naive approach would require comparing every response against every other response, resulting in $\binom{G}{2}$ evaluations and a computational complexity of $O(G^2)$.
Writing-Zero introduces Bootstrapped Relative Policy Optimization (BRPO).
Instead of exhaustive pairwise comparisons, BRPO employs a bootstrapping mechanism where a single response $o_{\text{ref}}$ is randomly sampled from the current group $\{o_i\}_{i=1}^G$ to serve as a dynamic anchor.
The quality of every other response $o_i$ is then evaluated relative to this anchor.
This strategy reduces the number of required GenRM calls from $O(G^2)$ to $O(G)$, maintaining the linear scalability characteristic of pointwise methods while retaining the superior signal quality of pairwise evaluation.
Subsequent work explores seeded single-elimination tournament ranking \citep{zhang2026arenarl} and alternating optimization of rubric generators and judges with a reference-anchor online RL setup \citep{xu2026alternating}.

\section{Approach}
\label{sec:approach}

In this section, we introduce the Open Rubric System (OpenRS), a transparent and structured rubric system based on the LLM-as-a-Judge paradigm (Figure~\ref{fig:framework}).
At the core of OpenRS is the \emph{meta-rubric}: a hierarchical guideline that does not evaluate responses directly, but instead specifies \emph{how} an unmodified LLM judge should generate evaluation criteria, assign weights, and handle failure modes.
During evaluation, the meta-rubric is instantiated into concrete, per-pair \textit{Adaptive Rubrics} for subjective assessment, complemented by \textit{Pointwise Verifiable Rubrics} that provide verifiable reward components and hard-constraint guardrails for objective sub-tasks with ground truth.
The meta-rubric itself is optimized offline via evolutionary refinement and human-in-the-loop domain adaptation (right side of Figure~\ref{fig:framework}).
This composite reward signal provides a robust foundation for consistent evaluation and scalable reinforcement learning in general domains.

\subsection{Overall Framework}
\label{sec:overall_framework_of_rubric_system}

We posit that a robust reward signal should jointly capture (i) relative preference over subjective dimensions (e.g., helpfulness and creativity), and (ii) absolute performance on objective, checkable dimensions (e.g., math correctness, strict formatting constraints).
Accordingly, our system decomposes the reward $R_i$ for an output $o_i$ given query $q$ into two complementary pathways: a pairwise adaptive pathway for subjective judgment and a pointwise verifiable pathway for objective supervision.

\paragraph{Hierarchical Meta Rubric.}
The foundation of OpenRS is the \textit{Meta Rubric} $\mathcal{M}$---a rubric-generation guideline (rather than a rubric applied to responses directly) that defines evaluation principles, criterion categories, weighting norms, and non-negotiable failure modes.
To balance generalization and specificity, we structure $\mathcal{M}$ hierarchically:
a \textit{General Meta Rubric} $\mathcal{M}_{\text{gen}}$ captures broadly applicable alignment principles and serves as the global foundation,
while \textit{Domain Meta Rubrics} $\mathcal{M}_{\text{dom}}$ extend it with fine-grained criteria tailored to specific domains (e.g., code generation, creative writing), i.e., $\mathcal{M} = \mathcal{M}_{\text{gen}} \oplus \mathcal{M}_{\text{dom}}$.
Crucially, $\mathcal{M}$ is optimized offline and remains fixed during evaluation---all adaptation happens at the principle level, not the model level, so the judge model itself is never trained.

\paragraph{Pairwise Adaptive Rubric.}
For subjective evaluation, we adopt a pairwise paradigm.
Because a static rubric cannot anticipate which dimensions matter for a particular comparison, given $\mathcal{M}$ OpenRS instantiates a dynamic \textit{Adaptive Rubric} $\mathcal{R}_{ij}$ for each response pair $(o_i, o_j)$.
Concretely, we first identify the salient semantic differences $\Delta_{ij}$, and then condition on them to contextualize $\mathcal{M}$ into a focused, pair-specific criterion set:
\begin{equation}
    \Delta_{ij} = f_{\text{diff}}(q, o_i, o_j), \quad \mathcal{R}_{ij} = f_{\text{adapt}}(\mathcal{M}, q, o_i, o_j, \Delta_{ij})
\end{equation}
where $f_{\text{diff}}$ extracts the semantic differences $\Delta_{ij}$ between the two responses $(o_i, o_j)$, and $f_{\text{adapt}}$ turns the Meta Rubric $\mathcal{M}$ into an Adaptive Rubric $\mathcal{R}_{ij}$, which contains a set of weighted criteria $\mathcal{R}_{ij} = \{(c_k, w_k)\}_{k=1}^K$.
Instead of predicting a single holistic preference probability, we evaluate the pair against each specific criterion.
For each criterion $c_k$, the model assigns a comparative score $v_k \in \{-2, -1, 0, 1, 2\}$, where negative values indicate $o_i$ is inferior to $o_j$, positive values indicate $o_i$ is superior, and 0 implies equality.
The criterion-level comparative verdicts are then \emph{externally} aggregated into a pairwise preference score $s_{ij}$, i.e., a weight-normalized average of the criterion-specific verdicts:
\begin{equation}
    s_{ij} = \frac{\sum_{k=1}^{K} w_k \cdot v_k}{\sum_{k=1}^{K} w_k}
\label{eq:pairwise_adaptive_rubric_score}
\end{equation}
Because the aggregation happens outside the judge, no absolute quality score is ever required, and every criterion-level judgment remains inspectable.
In the RL loop, we obtain a per-sample scalar signal by comparing each rollout $o_i$ against a reference anchor $o_{\text{ref}}$ (Section~\ref{sec:pairwise_genrm_brpo}), yielding the criterion-weighted score $s_{i,\text{ref}}$.

\paragraph{Pointwise Verifiable Rubric.}
Certain dimensions of evaluation are absolute and can be verified on a single output without pairwise comparison (e.g.,  word count, length or format constraints, and strict instruction following).
We capture these signals with a \textit{Pointwise Verifiable Rubric} $\mathcal{V}_q$, i.e., a set of verifiers derived from the query $q$.
Each verifier $c \in \mathcal{V}_q$ produces a deterministic signal $\phi_c(o)$ on a candidate output $o$ (e.g., pass/fail, exact match, or a bounded score), which can be used either as a hard gate or as an additive reward/penalty term.
This pointwise pathway therefore plays a dual role: it provides guardrails against degenerate behaviors and supplies verifiable reward components whenever ground-truth or programmatic checks are available.

\paragraph{Meta-Rubric Refinement.}
The quality of the reward signal depends critically on $\mathcal{M}$.
We therefore treat rubric design as a discrete optimization problem, aiming to maximize alignment with human preferences, and refine $\mathcal{M}$ at two levels:
an automated evolutionary beam-style search over criterion-level edits for $\mathcal{M}_{\text{gen}}$ (Section~\ref{sec:general_meta_rubric_refinement}), which uses a black-box mutation strategy to promote robustness,
and a data-aware, human-in-the-loop workflow for $\mathcal{M}_{\text{dom}}$ (Section~\ref{sec:domain_meta_rubric_refinement}), which leverages error analysis on preference data to perform targeted updates.

\begin{algorithm}[t]
  \caption{General Meta Rubric Refinement}
  \label{alg:general_meta_rubric_refinement}
  \KwIn{Seed rubric $\mathcal{M}_{\text{seed}}$; beam size $B$; total rollouts per iteration $G$; iterations $T$; oracle $R(\cdot)$; refinement policy $\pi_\theta$; reference policy $\pi_{\text{ref}}$; GRPO hyperparams $\varepsilon,\beta$}
  \KwOut{Refined general meta rubric $\mathcal{M}_{\text{gen}}$ and trained refinement policy $\pi_\theta$}

  Initialize beam $\mathcal{B}_0 \leftarrow \{\mathcal{M}_{\text{seed}}\}$ and fill to size $B$ by copying\;
  \For{$t \leftarrow 0$ \KwTo $T-1$}{
      Initialize candidate set $\mathcal{C} \leftarrow \emptyset$\;
      $g_{\text{per}} \leftarrow G / B$\;
      \ForEach{$\mathcal{M} \in \mathcal{B}_t$}{
          \For{$j \leftarrow 1$ \KwTo $g_{\text{per}}$}{
              Roll out an edit sequence $o \sim \pi_{\theta_{\text{old}}}(\cdot \mid \mathcal{M})$ with actions in $\mathcal{A}=\{\texttt{ADD},\texttt{DELETE},\texttt{MODIFY}\}$\;
              Apply edits to obtain mutated rubric $\tilde{\mathcal{M}} \leftarrow \textsc{ApplyEdits}(\mathcal{M}, o)$\;
              Score reward $R \leftarrow R(\tilde{\mathcal{M}})$\;
              Add $(\tilde{\mathcal{M}}, o, R)$ to $\mathcal{C}$\;
          }
      }
      Update beam $\mathcal{B}_{t+1} \leftarrow \text{Top}_B\big(\{ \tilde{\mathcal{M}} : (\tilde{\mathcal{M}}, o, R)\in \mathcal{C} \},\ \text{by }R\big)$\;

      Compute GRPO advantages $\{\hat{A}_{i,\tau}\}$ from rewards $\{R_i\}_{i=1}^{G}$ as in Eq.~\ref{grpo_advantage}\;
      Let $\mathcal{I} \leftarrow \text{Top}_B(\{R_i\}_{i=1}^{G})$ be indices of top-$B$ rollouts in the group\;
      Update $\pi_\theta$ by maximizing the Asym-GRPO objective over $i \in \mathcal{I}$:
      \[
        \frac{1}{B}\sum_{i \in \mathcal{I}} \frac{1}{|o_i|}\sum_{\tau=1}^{|o_i|}
        \Big(\min(r_{i,\tau}(\theta)\hat{A}_{i,\tau},\ \text{clip}(r_{i,\tau}(\theta),1-\varepsilon,1+\varepsilon)\hat{A}_{i,\tau})
        - \beta D_{\text{KL}}(\pi_\theta \| \pi_{\text{ref}})\Big)
      \]
      where $r_{i,\tau}(\theta)=\frac{\pi_\theta(o_{i,\tau}\mid \mathcal{M}_i, o_{i,<\tau})}{\pi_{\theta_{\text{old}}}(o_{i,\tau}\mid \mathcal{M}_i, o_{i,<\tau})}$\;
  }
  Return $\mathcal{M}_{\text{gen}} \leftarrow \arg\max_{\mathcal{M}\in \mathcal{B}_T} R(\mathcal{M})$ and $\pi_\theta$\;
\end{algorithm}

\subsection{General Meta Rubric Refinement}
\label{sec:general_meta_rubric_refinement}

We cast the refinement of the General Meta Rubric $\mathcal{M}_{\text{gen}}$ as a discrete optimization problem that aims to discover a robust,
constitution-level evaluation standard via a beam-style discrete search over criterion-level edits (Alg.~\ref{alg:general_meta_rubric_refinement}).
To promote domain generality and reduce overfitting, we adopt a black-box setting \emph{with respect to the evaluation data}:
the refinement policy $\pi_{\text{refine}}$ does not observe the preference dataset itself and only receives an aggregate scalar feedback from an evaluation oracle.
Concretely, given a candidate rubric $\mathcal{M}$, we use it to judge all pairs in a held-out preference set ($\sim$2K balanced, multi-category examples with human-annotated labels), and the evaluation oracle returns the deterministic pairwise accuracy $R(\mathcal{M})$---i.e., the fraction of pairs for which the rubric-based judgment agrees with the human preference.
Importantly, this oracle preference set is disjoint from all four evaluation benchmarks used in Section~\ref{sec:experiments}, so refinement cannot leak benchmark-specific information.

We initialize $\mathcal{M}_{\text{gen}}$ with an expert-crafted seed rubric and perform an evolutionary search augmented with beam selection.
At iteration $t$, we maintain a beam $\mathcal{B}_t=\{\mathcal{M}^{(b)}_t\}_{b=1}^{B}$ of the top-$B$ rubrics.
For each beam element $\mathcal{M}\in\mathcal{B}_t$, we roll out $g_{\text{per}}=G/B$ edit sequences $o \sim \pi_{\theta_{\text{old}}}(\cdot\mid \mathcal{M})$, where each edit action is drawn from $\mathcal{A}=\{\texttt{ADD},\texttt{DELETE},\texttt{MODIFY}\}$.
Applying $o$ to $\mathcal{M}$ yields a mutated rubric $\tilde{\mathcal{M}}=\textsc{ApplyEdits}(\mathcal{M},o)$, which is scored by $R(\tilde{\mathcal{M}})$.
After evaluating all $G$ candidates, we set the next beam $\mathcal{B}_{t+1}$ to the $\text{Top}_B$ candidates by reward.

Beam selection alone is insufficient when $\pi_{\text{refine}}$ is not trained, because the candidate pool is dominated by low-quality (often degrading) mutations.
We therefore train $\pi_{\text{refine}}$ so that its rollouts increasingly concentrate on edits that yield high-reward rubrics, making the beam search consistently improve over iterations.
We optimize $\pi_{\text{refine}}$ using a GRPO-style objective,
where at each iteration we treat the $G$ rollouts generated from $\mathcal{B}_t$ as one GRPO group: we compute group-wise advantages from the rewards $\{R_i\}_{i=1}^{G}$ as in Eq.~\ref{grpo_advantage}.
Since rewards are sparse and skewed, standard GRPO may assign positive advantages to ``less-bad'' failures under group-wise normalization (Eq.~\ref{grpo_advantage}).
To address this, we propose Asymmetric GRPO (Asym-GRPO), which perform an asymmetric update that backpropagates only through the $\text{Top}_B(\{R_i\}_{i=1}^{G})$ rollouts while masking all others.
This hard filtering concentrates learning on successful edits, avoids reinforcing ``less-bad'' failures under skewed rewards, and stabilizes improvement across iterations.
In our implementation, $\pi_{\text{refine}}$ is instantiated with Qwen3-30B-A3B, while the oracle judge that scores each candidate rubric is Qwen3-235B-A22B-Instruct-2507.
As shown in Table~\ref{tab:ablation_study_of_openrs}, replacing the refined $\mathcal{M}_{\text{gen}}$ with the initial $\mathcal{M}_{\text{seed}}$ (while keeping the evaluation prompt unchanged) causes a substantial performance drop ($85.4$ vs.\ $90.7$ on RewardBench v2), confirming that evolutionary refinement discovers genuinely stronger evaluation principles rather than merely a better prompt format.

\subsection{Domain Meta Rubric Refinement}
\label{sec:domain_meta_rubric_refinement}

While the General Meta Rubric captures broadly applicable alignment principles, many domains (e.g., code generation and creative writing) require additional, domain-specific criteria.
In particular, evaluation settings with heterogeneous subjective preferences (e.g., the PPE Chinese subset spanning 40+ user-facing categories) benefit most from domain-specific adaptation, whereas benchmarks with strong, largely objective norms are already well covered by $\mathcal{M}_{\text{gen}}$ alone (Appendix~\ref{sec:domain_rubric_analysis}).
We construct the Domain Meta Rubric $\mathcal{M}_{\text{dom}}$ by adapting the refined $\mathcal{M}_{\text{gen}}$ using a small, curated set of in-domain preference data.
These data are used for aggregate error analysis and domain-level principle edits, \emph{not} as input to per-pair adaptive-rubric instantiation, so the instantiated criteria remain grounded in the observable differences of each pair.
Unlike the black-box exploration used for $\mathcal{M}_{\text{gen}}$, domain adaptation is carried out in a data-aware manner via error analysis rather than gradient-based reinforcement learning.
Concretely, we compare rubric-based judgments against human labels, identify systematic failure modes, and apply targeted criterion-level edits from $\{ \texttt{ADD}, \texttt{DELETE}, \texttt{MODIFY} \}$.
To mitigate overfitting, we explicitly constrain updates to introduce abstract,
reusable principles (instead of case-specific heuristics) and validate the adapted rubric on held-out domain splits to ensure it remains generalizable and free of dataset artifacts.
Importantly, this is a human-in-the-loop workflow: human experts periodically audit proposed edits, resolve ambiguous failure cases, and approve revisions before merging them into $\mathcal{M}_{\text{dom}}$,
enabling rapid specialization without training an additional proposer policy.
This design is also motivated by cost: training the general refinement policy already requires substantial compute (see Table~\ref{tab:rl_training_config}); replicating it for every domain would multiply the cost linearly with the number of domains, whereas the human-in-the-loop workflow avoids this overhead entirely.

\subsection{Applying OpenRS to Reinforcement Learning}
\label{sec:openrs_rl}

We apply the Open Rubric System (OpenRS) as a reward interface for reinforcement learning on open-ended generation.
In contrast to RLHF \citep{ouyang2022training}, which depends on extensive human annotations, and RLVR \citep{guo2025deepseek},
which is largely limited to domains with deterministic ground truth,
OpenRS provides a scalable reward signal by converting \emph{evaluation quality} into training feedback:
we continuously refine the hierarchical Meta Rubric to improve the fidelity of the judge, which in turn improves the reward used for policy optimization.

To integrate our pairwise adaptive rubric into the RL loop efficiently, we adapt Bootstrapped Relative Policy Optimization (BRPO; Section~\ref{sec:pairwise_genrm_brpo}) for effective rubric-based pairwise comparisons.
For a rollout $o_i$, OpenRS constructs the final scalar reward as a superposition of a \textit{Pairwise Adaptive Score} and a \textit{Pointwise Verifiable Score}:
\begin{equation}
    R(q, o_i) = s_{i, \text{ref}} + \gamma \sum_{c \in \mathcal{V}_q} \phi_c(o_i),
\end{equation}
where $s_{i, \text{ref}}$ is the criterion-weighted preference score produced by the Adaptive Rubric $\mathcal{R}_{i, \text{ref}}$,
and the second term aggregates verifiable constraint signals from the Pointwise Verifiable Rubric $\mathcal{V}_q$.
For each constraint $c \in \mathcal{V}_q$, the function $\phi_c(o_i)$ returns $+1$ if the response $o_i$ satisfies condition $c$, and $-1$ otherwise.
The coefficient $\gamma$ serves as a balancing hyperparameter, ensuring that hard constraints impose a significant influence on the reward landscape.
This decomposition decouples subjective quality optimization from objective constraint enforcement, enabling stable RL updates under a transparent, rubric-based reward.

\begin{table}[t]
\centering
\resizebox{0.99\textwidth}{!}{%
\begin{tabular}{lccccccc}
\toprule
\textbf{Model}                           & \textbf{RewardBench v2} & \textbf{PPE Pref / ZH} & \textbf{RM-Bench} & \textbf{JudgeBench} & \textbf{Avg.} \\
\midrule

\noalign{\vskip 1ex}
\rowcolor{gray!20} \multicolumn{6}{c}{\textit{LLM-as-a-Judge \& Generative Reward Models \& Rubric}} \\
\noalign{\vskip 1ex}
DeepSeek-GRM-27B \citep{liu2025inference}                                                                         &  -   & 71.1\textsuperscript{*} &  -   &  -   & - \\
DeepSeek-GRM-27B (w/ MetaRM) \citep{liu2025inference}                                                             &  -   & 72.2\textsuperscript{*} &  -   &  -   & - \\
RM-R1-Qwen-Instruct-32B \citep{chen2025rm}                                                                        &  -   & 76.3\textsuperscript{*} & 79.1 &  -   & - \\
RM-R1-DeepSeek-Distill-Qwen-32B \citep{chen2025rm}                                                                &  -   & 75.3\textsuperscript{*} & 83.9 &  -   & - \\
EvalPlanner (Llama-3.3-70B) \citep{saha2025learning}                                                              &  -   &  -                      & 82.1 & 56.6 & - \\
J1-Llama-70B \citep{whitehouse2025j1}                                                                             &  -   &  -                      & 82.7 & 60.0 & - \\
RationaleRM (Qwen3-30B-A3B) \citep{wang2026outcomeaccuracyenoughaligning}                                         &  -   &  -                      & 87.1 & 82.0 & - \\
Auto-Rubric (UltraFeedback, Qwen3-235B-A22B-Instruct-2507) \citep{xie2025auto}  &  -                      & 78.0\textsuperscript{*} &  -                      & 74.3\textsuperscript{*} &  -                      \\
\midrule

\noalign{\vskip 1ex}
\rowcolor{gray!20} \multicolumn{6}{c}{\textit{Open Scalar Reward Models}} \\
\noalign{\vskip 1ex}
Llama-3-OffsetBias-RM-8B \citep{park2024offsetbias}                                                               & 64.8 & 60.6\textsuperscript{*} & 71.3 & 63.5 & 65.0 \\
ArmoRM-Llama3-8B-v0.1 \citep{wang2024interpretable}                                                               & 66.5 & 58.3\textsuperscript{*} & 69.3 & 59.7 & 63.4 \\
Internlm2-20b-reward \citep{cai2024internlm2}                                                                     & 56.3 & 69.4\textsuperscript{*} & 68.3 & 64.3 & 64.8 \\
LDL-Reward-Gemma-2-27B-v0.1                                                                                       & 72.5 & 43.3\textsuperscript{*} & 71.1 & 64.2 & 67.4 \\
Llama-3.1-Nemotron-70B \citep{wang2024helpsteer2preference}                                                       & 76.7 & 68.7\textsuperscript{*} & 72.2 & 65.8 & 70.5 \\
INF-ORM-Llama3.1-70B \citep{INF-ORM-Llama3.1-70B}                                                                 & 76.5 & 65.7\textsuperscript{*} & 73.8 & 70.2 & 71.7 \\
Skywork-Reward-Llama-3.1-8B-v0.2 \citep{liu2024skywork}                                                           & 71.8 & 62.2\textsuperscript{*} & 72.1 & 62.9 & 67.1 \\
Skywork-Reward-Gemma-2-27B-v0.2 \citep{liu2024skywork}                                                            & 75.3 & 57.6\textsuperscript{*} & 70.0 & 66.5 & 67.3 \\
Skywork-Reward-V2-Qwen3-8B \citep{liu2025skyworkrewardv2scalingpreferencedata}                                    & 78.2 & 76.4\textsuperscript{*} & 82.6 & 73.4 & 77.6 \\
Skywork-Reward-V2-Llama-3.1-8B \citep{liu2025skyworkrewardv2scalingpreferencedata}                                & 84.1 & 79.6\textsuperscript{*} & 92.8 & 80.0 & 84.3 \\
\midrule

\noalign{\vskip 1ex}
\rowcolor{gray!20} \multicolumn{6}{c}{\textit{LLM-as-a-Judge \& Rubric System}} \\
\noalign{\vskip 1ex}
OpenRS (Qwen3-8B)                                                                                                 & 75.1          & 73.5          & 92.4          & 89.5          & 82.6 \\
OpenRS (gpt-oss-120b)                                                                                             & 85.7          & 76.6          & 91.7          & \textbf{93.3} & 86.8 \\
OpenRS (DeepSeek-V3.1)                                                                                            & 85.6          & 78.1          & 91.8          & 89.1          & 86.2 \\
OpenRS (DeepSeek-V3.2)                                                                                            & 85.8          & 78.9          & 92.7          & 89.4          & 86.7 \\
OpenRS (Qwen3-30B-A3B-Instruct-2507)                                                                              & 83.7          & 78.4          & 91.1          & 86.4          & 84.9 \\
OpenRS (Qwen3-235B-A22B-Instruct-2507)                                                                            & \textbf{90.7} & \textbf{82.3} & \textbf{93.0} & 91.6          & \textbf{89.4} \\
\bottomrule
\end{tabular}}
\vspace{-0.5em}
\caption{Main benchmark comparison of OpenRS with scalar reward models (SRMs), generative judges, and rubric-based methods.
  For OpenRS, parentheses denote the \emph{unmodified} LLM judge backbone.
  \textbf{Bold} numbers indicate the best performance among all models.
  Results marked with "*" are re-implemented.
  Entries marked with "-" indicate either (1) that the specific model's score is not included in the original benchmark or (2) that the model is unreleased.
}
\label{tab:key_reward_model_performance}
\end{table}

\section{Experiments}
\label{sec:experiments}
Most RL pipelines for general-purpose LMs rely on scalar reward models (SRMs),
which compress multi-dimensional human preferences into a single score and often induce reward hacking and brittle generalization.
In contrast, OpenRS provides a transparent,
rubric-based reward signal that decomposes subjective preference into interpretable criteria while optionally incorporating verifiable constraint signals.
Before deploying OpenRS as a scalable reward interface for RL,
we show that its rubric-based judgments consistently outperform strong SRM baselines in alignment with human preferences on standard reward-modeling benchmarks.

\subsection{Experimental Setup}
We evaluate OpenRS against state-of-the-art SRMs on four diverse benchmarks: \textbf{RM-Bench} \citep{liu2024rmbenchbenchmarkingrewardmodels}, \textbf{JudgeBench} \citep{tan2025judgebenchbenchmarkevaluatingllmbased}, \textbf{RewardBench v2} \citep{malik2025rewardbench2advancingreward}, and \textbf{PPE Preference} \citep{frick2024evaluaterewardmodelsrlhf}.
For the rubric refinement process in Sec~\ref{sec:general_meta_rubric_refinement}, the refinement policy $\pi_{\text{refine}}$ is Qwen3-30B-A3B and oracle scoring uses Qwen3-235B-A22B-Instruct-2507\footnote{https://huggingface.co/Qwen/Qwen3-235B-A22B-Instruct-2507}, with beam size $B=4$ and $G=32$ rollouts per iteration.
We use the resulting refined meta-rubric throughout; its oracle preference set is disjoint from all four evaluation benchmarks, preventing benchmark leakage.
PPE Preference is multilingual ($\sim$16k examples) and includes a Chinese subset ($\sim$1.2k); we report results on the Chinese subset and mark obvious label errors as \texttt{label\_error} after cross-checking with multiple strong LLM judges and human review.
To mitigate position bias, all pairwise evaluations are conducted bidirectionally: we evaluate both $(o_i, o_j)$ and $(o_j, o_i)$ and accept a preference only when both directions agree; contradictory judgments are labeled ``Same'' (see Section~\ref{sec:same_rate}).
For reproducibility, we release our refined Meta Rubric, Verifiable Rubric, the complete evaluation scripts, and all preference data for these four benchmarks at \url{https://github.com/Qwen-Applications/OpenRS}.

\subsection{Main Results}
As shown in Table \ref{tab:key_reward_model_performance},
OpenRS achieves state-of-the-art alignment with human preferences across all four benchmarks and attains the best overall average.
Notably, OpenRS (Qwen3-235B-A22B) outperforms the strongest open SRM baseline (Skywork-Reward-V2) by $+5.1$ points on the average score ($89.4$ vs.\ $84.3$),
indicating that explicit rubric decomposition provides a strictly stronger training signal than monolithic scalar scoring.
OpenRS also outperforms all LLM-based judge baselines that require domain-specific training (top block of Table~\ref{tab:key_reward_model_performance}): RM-R1-32B \citep{chen2025rm} ($76.3$* on PPE), J1-70B \citep{whitehouse2025j1} ($60.0$ on JudgeBench), and Auto-Rubric \citep{xie2025auto} ($78.0$*/$74.3$* on PPE/JudgeBench, re-evaluated with the same 235B backbone using a pointwise rubric).
Crucially, the judge model used for benchmark evaluation is entirely unmodified---only the meta-rubric, a plain text artifact, is optimized via a separate refinement policy---confirming that the structured pipeline, rather than specialized judge training, is the key differentiator.

RM-Bench probes sensitivity to subtle content differences and robustness to style/verbosity biases (the substantively better answer may be shorter or less formatted);
OpenRS (Qwen3-235B-A22B) achieves $93.0$, surpassing the strongest open SRM baseline ($92.8$) by $+0.2$,
as criterion-level decomposition prevents superficial attributes from dominating the judgment.

JudgeBench targets judge reliability under common biases (position bias, verbosity bias, and self-preference) on complex reasoning; OpenRS (gpt-oss-120b) reaches $93.3$, improving over the strongest reported baseline ($82.0$) by $+11.3$, consistent with our bias-aware meta-rubric principles and pair-difference-grounded adaptive comparisons.

RewardBench v2 covers a broad mix of capabilities (Chat/Reasoning/Safety) with challenging, structured, out-of-distribution preference pairs; OpenRS (Qwen3-235B-A22B) attains $90.7$, exceeding the strongest open SRM baseline ($84.1$) by $+6.6$, as adaptive rubric generation dynamically selects and weights criteria while the meta-rubric provides transferable high-level principles.

PPE Preference evaluates preference prediction on policy-generated outputs, better reflecting the on-policy distribution encountered during RLHF (we report the Chinese subset); OpenRS (Qwen3-235B-A22B) achieves $82.3$, outperforming the strongest open SRM baseline ($79.6$) by $+2.7$, supported by higher-fidelity rubric-based comparisons and our beam-search refinement with top-$B$ filtered GRPO updates.

\paragraph{Different Foundation Models for OpenRS}
The primary objective of our Rubric System is to serve as a robust reward engine for Reinforcement Learning.
The RL training process inherently involves generating a massive volume of rollouts, which necessitates a reward signal that is not only high-quality but also computationally accessible.
Relying on commercial, closed-source APIs for this purpose is impractical due to prohibitive costs and rate limits.
Consequently, our system is designed to be deployed locally, depending entirely on the capabilities of open-weights models to provide scalable feedback APIs.
Finally, we analyze OpenRS under different open-weights foundation models as the judge backbone.
The results show a clear scaling trend: stronger backbones yield stronger rubric evaluation (e.g., OpenRS improves from $84.9$ with Qwen3-30B-A3B to $89.4$ with Qwen3-235B-A22B on the average score), indicating that evaluation quality in OpenRS scales with the underlying model's reasoning and instruction-following ability.
Moreover, OpenRS powered by non-Qwen backbones such as gpt-oss-120b also surpasses strong SRM baselines on key benchmarks (e.g., JudgeBench), while DeepSeek variants remain competitive where scores are available, supporting the generality of our rubric-based approach beyond a single model family.

Importantly, the benefit of the pipeline is not restricted to very large judges.
Even with an \emph{unmodified} Qwen3-8B as the judge, OpenRS achieves $92.4$ on RM-Bench and $89.5$ on JudgeBench---essentially matching the 235B configuration ($93.0$/$91.6$) and far exceeding the strongest SRM baseline ($92.8$/$82.0$).
The remaining RewardBench v2 gap ($75.1$ vs.\ $90.7$) traces mainly to the Precise IF subset ($26.9\%$; see Table~\ref{tab:rm_bench_and_rewardbench_detailed_scores}), where verifying fine-grained instruction-following demands capabilities that scale with model size;
on Math ($92.9\%$), Safety ($88.5\%$), and Ties ($83.9\%$) the 8B backbone remains competitive, confirming that the pipeline is effective across scales and that residual gaps reflect backbone limitations rather than deficiencies of OpenRS itself.
A controlled same-backbone comparison makes this especially clear: against Skywork-Reward-V2-Qwen3-8B (identical Qwen3-8B backbone, trained on 26M preference pairs), OpenRS (Qwen3-8B) averages $82.6$ vs.\ $77.6$ ($+5.0$) across all four benchmarks, with $+9.8$ on RM-Bench and $+16.1$ on JudgeBench---without any judge training at all.
Similarly, with Qwen3-30B-A3B (only 3B active parameters) OpenRS reaches $84.9$, already exceeding Skywork-Reward-V2-Llama-3.1-8B ($84.3$).

\paragraph{Pareto Frontier of OpenRS}

\begin{figure}[t]
    \centering
    \includegraphics[width=0.8\textwidth]{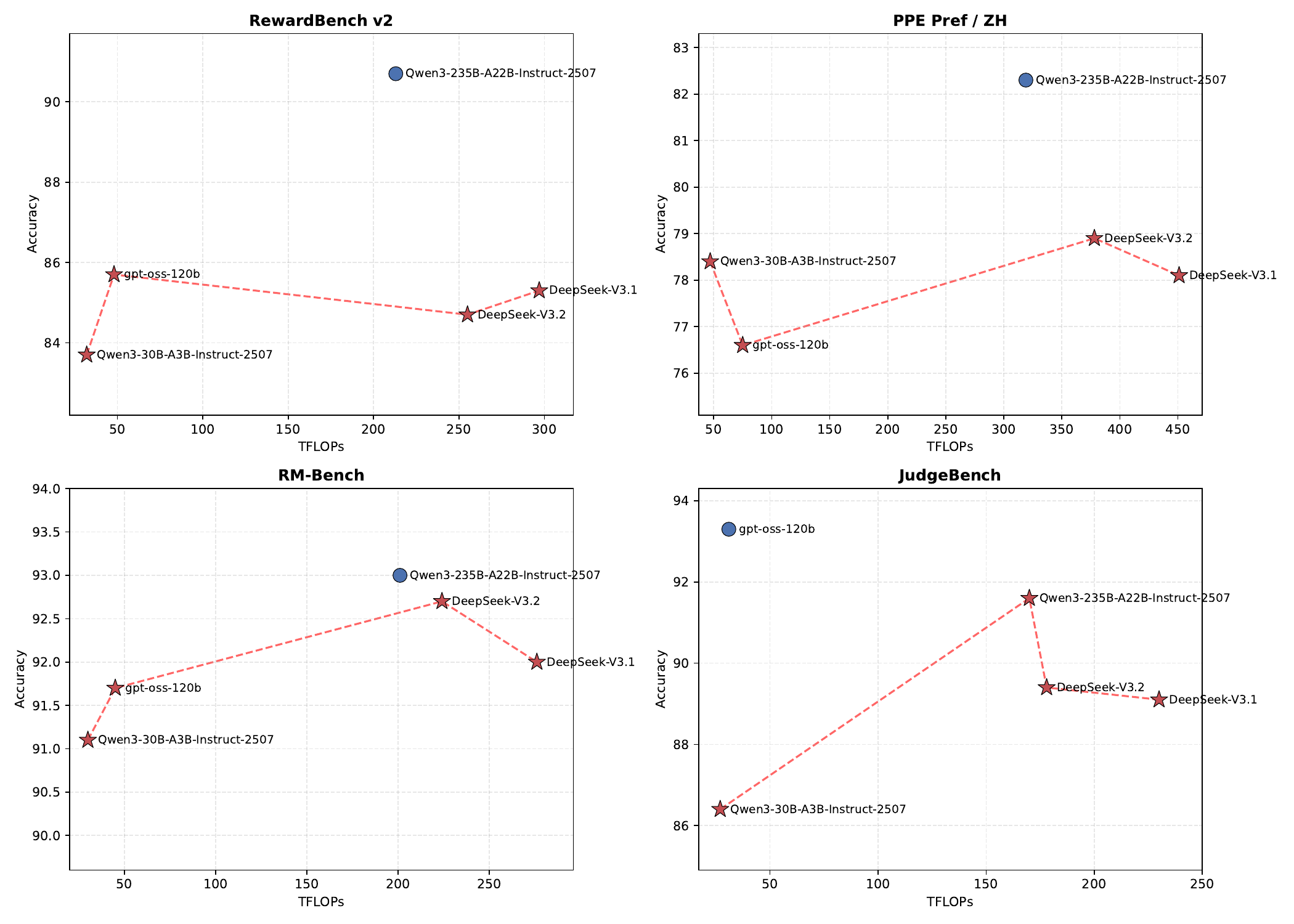}
    \captionsetup{justification=centering}
    \caption{Pareto Frontier for Open Rubric System on Different Benchmarks.}
    \label{fig:pareto_frontier}
\end{figure}

Figure~\ref{fig:pareto_frontier} further characterizes this trade-off by plotting OpenRS accuracy against the judge backbone's inference compute (FLOPs).
Across benchmarks, Qwen3-Instruct models form a strong Pareto frontier: Qwen3-235B-A22B achieves the best overall evaluation quality while remaining substantially more compute-efficient than heavier backbones that do not improve accuracy (e.g., DeepSeek variants on several tasks).
Although gpt-oss-120b can be particularly strong on JudgeBench, Qwen3-235B-A22B is the most consistently competitive choice across all evaluated benchmarks.
Given that RL requires both high-fidelity rewards and high-throughput scoring, this Pareto analysis motivates our default choice of Qwen3-235B-A22B-Instruct-2507 as the judge backbone for subsequent RL training.

\section{Analysis}

\begin{table}[t]
\centering
\resizebox{0.80\textwidth}{!}{%
\begin{tabular}{lccccccc}
\toprule
\textbf{Model}                           & \textbf{RewardBench v2} & \textbf{PPE Pref / ZH} & \textbf{RM-Bench} & \textbf{JudgeBench} & \textbf{Avg.} \\
\midrule
Pointwise OpenRS                         & 86.1 & 76.2          & 91.1          & 90.9          & 86.1 \\
Two-stage (query-only)                   & 86.2 & 75.5          & -             &  -            & - \\
Seed rubric (same prompt)                & 85.4 & 74.6          & -             &  -            & - \\
OpenRS                                   & \textbf{90.7} & \textbf{82.3} & \textbf{93.0} & 91.6          & \textbf{89.4} \\
\,\,\, - w/o diff mechanism              & 88.9 & 80.6          & 91.8          & 88.5          & 87.5 \\
\,\,\, - w/o domain meta rubric          &  -   & 80.9          & -             &  -            & - \\
\bottomrule
\end{tabular}}
\vspace{-0.5em}
\captionsetup{justification=centering}
\caption{Ablation study of OpenRS based on Qwen3-235B-A22B-Instruct-2507.}
\label{tab:ablation_study_of_openrs}
\end{table}

\subsection{Recipe for OpenRS}
\label{sec:recipe_for_openrs}

In this subsection, we summarize practical design choices for building an effective rubric system.
Rather than treating these components as independent add-ons, we recommend using them as a coherent ``recipe'' that jointly improves evaluation fidelity and robustness across benchmarks.

\paragraph{Refine the meta rubric, not just the prompt.}
Before examining the pairwise/pointwise choice, we first verify that the gains come from \emph{optimized principles} rather than from prompt engineering.
Replacing the refined $\mathcal{M}_{\text{gen}}$ with the initial seed rubric while keeping the evaluation prompt byte-identical (Table~\ref{tab:ablation_study_of_openrs}, row ``Seed rubric (same prompt)'') causes a large drop ($85.4$ vs.\ $90.7$ on RewardBench v2 and $74.6$ vs.\ $82.3$ on PPE Pref/ZH).
This confirms that evolutionary refinement discovers substantively stronger evaluation principles, and that the meta-rubric---not the surrounding scaffold---is the primary carrier of evaluation quality in OpenRS.

\paragraph{Pairwise over Pointwise.}
OpenRS is designed to be flexible: the hierarchical Meta Rubric and the dynamic Adaptive Rubric can be instantiated under either pairwise comparison or pointwise scoring.
In practice, pairwise evaluation is a better default for capturing nuanced preferences (Section~\ref{sec:pairwise_genrm_brpo}),
while a pointwise variant can be useful when pairwise comparisons are infeasible.
Our ablation in Table~\ref{tab:ablation_study_of_openrs} (row ``Pointwise OpenRS'') shows a clear performance gap to full OpenRS, supporting pairwise comparison as the recommended configuration.

\paragraph{Think differences first to ground criterion selection.}
When generating an Adaptive Rubric for a response pair $(o_i,o_j)$, we first identify the salient semantic differences $\Delta_{ij}$ and then synthesize criteria conditioned on these differences (Section~\ref{sec:overall_framework_of_rubric_system}).
This ``diff-first'' step helps prevent generic, underspecified criteria and reduces susceptibility to superficial biases by anchoring the evaluation to what actually distinguishes the two candidates.
We ablate this design at two granularities.
First, a \textit{Two-stage} baseline that generates the rubric from the query alone (never seeing the two responses before criterion selection) drops to $86.2$/$75.5$ on RewardBench v2 and PPE Pref/ZH.
Second, even keeping both responses visible but skipping the explicit extraction of $\Delta_{ij}$ (Table~\ref{tab:ablation_study_of_openrs}, row ``w/o diff mechanism'') still degrades performance to $88.9$/$80.6$.
Together these results indicate that \emph{structured} difference analysis---not merely response visibility---is an essential ingredient of OpenRS.

\paragraph{Use Domain Meta Rubrics to scale reliably across domains.}
A key advantage of OpenRS is that it decouples principle specification from model parameters:
to specialize the reward signal for a target domain, we refine or introduce a \emph{Domain Meta Rubric} rather than fine-tuning a generative reward model.
This modularity avoids catastrophic forgetting and negative transfer that can occur when a single monolithic reward model is adapted across heterogeneous domains.
The ablation in Table~\ref{tab:ablation_study_of_openrs} (row ``w/o domain meta rubric'') confirms that domain-specific refinements matter most where subjective preferences are heterogeneous: on PPE (ZH), which spans 40+ user-facing categories, adding $\mathcal{M}_{\text{dom}}$ improves accuracy from $80.9$ to $82.3$.
Benchmarks with strong, largely objective norms are already well covered by $\mathcal{M}_{\text{gen}}$ alone; we analyze when domain adaptation is (and is not) needed in Appendix~\ref{sec:domain_rubric_analysis}.

\subsection{Refinement Policy: Frozen vs. GRPO vs. Asym-GRPO}

\begin{figure}[t]
    \centering
    \includegraphics[width=0.5\textwidth]{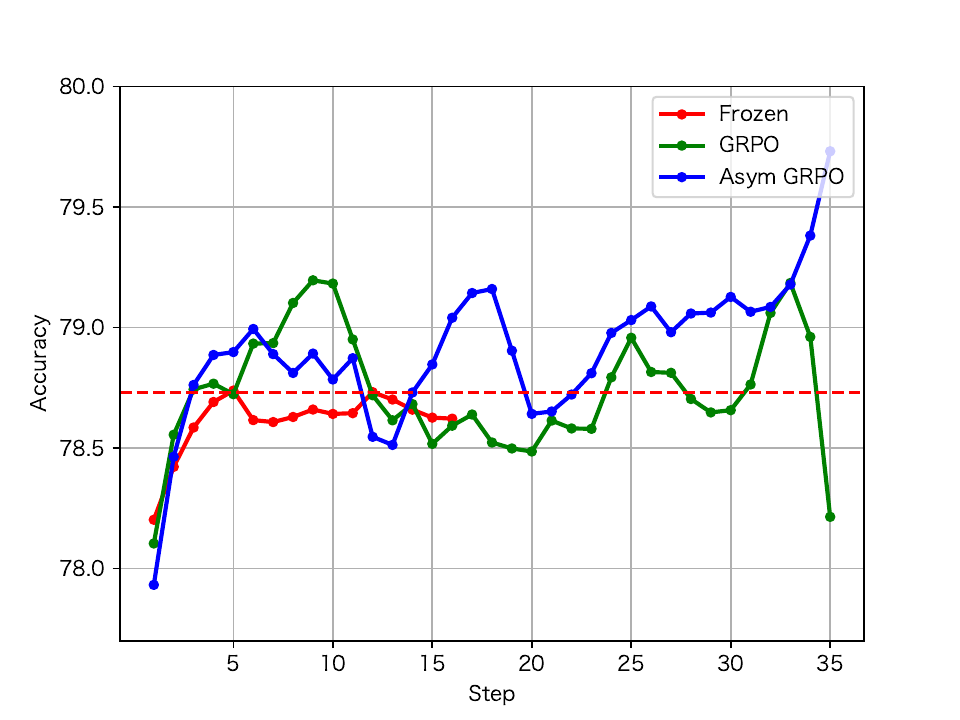}
    \caption{Training dynamics of the refinement policy $\pi_{\text{refine}}$ under different settings.}
    \label{fig:grpo_vs_asym_grpo}
\end{figure}

In Section \ref{sec:general_meta_rubric_refinement},
we argued that training the refinement policy $\pi_{\text{refine}}$ is crucial for effective discrete meta-rubric search,
and that naive GRPO updates can be unstable under sparse, skewed rewards.
We compare three settings: (i) \textbf{Frozen},
where $\pi_{\text{refine}}$ is kept fixed and only beam selection is performed; (ii) \textbf{GRPO},
which updates $\pi_{\text{refine}}$ using all rollouts with standard group-wise normalization;
(iii) \textbf{Asym-GRPO}, which masks gradients from rollouts outside $\text{Top}_B(\{R_i\}_{i=1}^{G})$.
As shown in Figure \ref{fig:grpo_vs_asym_grpo},
Frozen quickly plateaus, standard GRPO exhibits noticeable volatility,
whereas Asym-GRPO yields more stable and sustained improvements,
supporting its role in reinforcing successful mutations while avoiding ``less-bad'' failures.

\subsection{Say Goodbye to Scalar Reward Model: General RL with Open Rubric System}
\label{sec:general_rl_with_open_rubric_system}

In Section \ref{sec:openrs_rl}, we described how OpenRS can serve as a scalable reward interface for RL by combining pairwise adaptive comparisons with optional verifiable constraints.
Here, we provide an end-to-end RL comparison to test whether replacing an SRM with OpenRS yields measurable improvements during policy optimization.
To isolate reward quality, all RL settings are held fixed across variants---the same warm-start in-house policy model and the same RL training data---and the only change is replacing the scalar reward model used for scoring with OpenRS (powered by Qwen3-235B-A22B-Instruct-2507).
As shown in Table~\ref{tab:scalar_rm_and_rlpr}, this simple replacement improves the average performance on five public benchmarks, with particularly notable gains on instruction-following stress tests (IFEval/IFScale) and JudgeMark-v2.
We also include \textit{RL + Pointwise OpenRS} as an intermediate baseline: adopting the same meta-rubric principles but reverting to pointwise rubric scoring yields a modest improvement over the scalar RM, yet remains clearly behind the full pairwise OpenRS reward.
This gap supports our central claim that \emph{pairwise}, difference-grounded rubric execution and external preference aggregation are key for producing a high-fidelity training signal in open-ended RL, beyond the benefits of rubric \emph{formatting} alone.
Qualitatively, the OpenRS-trained model exhibits richer subjective expression and clearer stance-taking compared to the ``assistant-like'' neutrality of scalar-RM baselines, suggesting that preserving multi-dimensional preference structure leads to better alignment in open-ended generation (Appendix~\ref{sec:case_study}, Figs.~\ref{fig:case_1_zh}--\ref{fig:case_1_en}).
We further track two training-time diagnostics in Figure~\ref{fig:same_rate_and_entropy_during_rl}: policy entropy and the fraction of ``Same'' judgments produced by the pairwise adaptive rubric, which together provide a coarse view of exploration dynamics and judge decisiveness during RL.

\begin{table}[t]
\centering
\vspace{-0.5em}
\resizebox{0.9\textwidth}{!}{%
\begin{tabular}{lcccccc}
\toprule
\textbf{Model}                                & \textbf{IFEval}   & \textbf{IFScale} & \textbf{EQ-Bench} & \textbf{JudgeMark-v2} &  \textbf{Chinese Simple QA} & \textbf{Avg.} \\
\midrule
\noalign{\vskip 1ex}
RL + Scalar Reward Model                      & 82.3   & 54.8 & 82.2 & 43.8 & 78.9 & 68.4 \\
RL + Pointwise OpenRS                         & 82.7   & 56.2 & 81.6 & 49.4 & 77.2 & 69.4 \\
RL + OpenRS (Qwen3-235B-A22B-Instruct-2507)   & 83.0   & 59.0 & 81.2 & 54.4 & 79.1 & 71.3 \\
\bottomrule
\end{tabular}}
\vspace{-0.5em}
\caption{Performance comparison of RL tasks on public benchmarks with scalar reward model and our rubric system.}
\label{tab:scalar_rm_and_rlpr}
\end{table}

\paragraph{Training Efficiency}

Because RL produces a large volume of rollouts, a practical rubric-based reward must be both high-quality and throughput-efficient.
In our implementation, we deploy Qwen3-235B-A22B-Instruct-2507 as a local judge service on 128 GPUs (80GB each) and use the \texttt{verl} framework for training.
With asynchronous request batching and agent-loop overlap, the majority of judge calls can be pipelined with policy rollouts; in our stress tests, OpenRS sustains $\sim$10{,}000 concurrent requests under the most efficient configuration, resulting in only a small overhead on wall-clock training time compared to an SRM-based setup.
Table~\ref{tab:rl_training_config} provides the full training configuration and wall-clock breakdown for our end-to-end RL run.

\begin{table}[t]
\centering
\small
\setlength{\tabcolsep}{4pt}
\resizebox{0.55\textwidth}{!}{%
\begin{tabular}{ll}
\toprule
\textbf{Configuration} & \textbf{Value} \\
\midrule
\multicolumn{2}{l}{\textit{Model \& Infrastructure}} \\
Policy model         & 235B (MoE, 22B active) \\
Framework            & \texttt{verl} \\
Judge backbone       & Qwen3-235B-A22B \\
Judge GPUs           & 128$\times$80GB \\
Peak memory/GPU      & 71.7 GB \\
\midrule
\multicolumn{2}{l}{\textit{RL Training Statistics}} \\
Total steps          & 149 \\
Wall-clock time      & $\sim$123 h ($\sim$5.1 days) \\
Time per step        & $\sim$47.7 min \\
Throughput           & $\sim$37 tokens/s \\
MFU (actor)          & $\sim$21\% \\
\midrule
\multicolumn{2}{l}{\textit{Per-Step Time Breakdown}} \\
Rollout generation   & 930.8s (33\%) \\
Reward scoring (OpenRS) & 1280.2s (45\%) \\
Actor update         & 377.2s (13\%) \\
Other                & 274.5s (9\%) \\
\bottomrule
\end{tabular}}
\vspace{-0.5em}
\caption{RL training configuration and wall-clock breakdown for end-to-end policy optimization with OpenRS.}
\label{tab:rl_training_config}
\end{table}

\subsection{More Analysis}

We provide fine-grained breakdowns and additional diagnostics in the appendix (Section~\ref{sec:additional_analysis}):
(i) \textit{Detailed Performance Comparison} decomposes results by RM-Bench difficulty levels, RewardBench v2 categories, and JudgeBench subdomains, showing where OpenRS improves robustness and where different SRM/GenRM baselines fail;
(ii) \textit{Generalization on Arena} evaluates LMArena-style generalization on in-house QA/Writing test sets and we defer final conclusions until evaluation completes;
(iii) \textit{Same Rate of Open Rubric System} quantifies judge uncertainty (``Same'' decisions) and supports efficient filtering during RL without costly majority voting.
We additionally include qualitative case studies illustrating behavioral differences induced by OpenRS-based RL.

\subsection{Aha Moment in General Domains}
\label{sec:aha_moment}

The "Aha Moment" observed in reasoning tasks, such as the emergence of self-correction and long-chain reasoning in DeepSeek-R1 \citep{guo2025deepseek}, suggests that reinforcement learning can unlock capabilities beyond the supervision data.
However, observing similar phenomena in open-ended general domains has remained elusive. We argue that the traditional Scalar Reward Model (SRM) is the primary bottleneck.
The true distribution of human preferences is inherently complex and multimodal.
SRMs attempt to approximate this distribution by learning from pairwise preference data, which often suffers from limited expressiveness and noise patterns.
Consequently, the SRM tends to fit a "trivial distribution" that captures only surface-level features of high-quality responses.
Policy optimization then acts as a "zero-forcing" process, compelling the policy model to collapse onto the mode of this imperfect proxy distribution.
This "double approximation loss", first in reward modeling and second in policy fitting, severely constrains the policy's upper bound, preventing it from exploring the true high-reward regions of the human preference landscape.
DeepSeek-R1 circumvents this by using rule-based verification for math and code, eliminating the first approximation step and allowing the policy to interact directly with the ground truth, thereby triggering the Aha Moment.

We posit that open-ended tasks possess a similar potential for Aha Moments, provided the reward signal avoids a learned pointwise scalar proxy.
Our Rubric System functions as a "soft" rule-based verifier for general tasks.
By explicating evaluation criteria via the Meta Rubric and performing criterion-wise pairwise comparisons, it preserves comparative evidence before forming RL advantages, rather than compressing it into a single learned score.
Instead, it presents the policy with a high-dimensional, interpretable feedback landscape, preserving the complexity of human preferences.
This setup mirrors the direct verification in reasoning tasks, creating the conditions necessary for emergence.
We specifically examine the policy's behavior during training, focusing on the entropy curve shown in Figure \ref{fig:entropy_during_rl}.
Unlike standard RLHF where entropy typically decreases monotonically as the model collapses to the reward mode, our method shows a distinctive pattern: an initial decrease followed by a significant rise and stabilization at a higher level.

While Table \ref{tab:scalar_rm_and_rlpr} provides a macroscopic performance comparison, we conduct a microscopic analysis of response dynamics to evidence this emergence.
Comparing these responses with those from a Scalar RM baseline and the state-of-the-art Gemini-3 Pro, we observe a qualitative shift.
As detailed in the Case Study (Section \ref{sec:case_study}), specifically Figures \ref{fig:case_1_zh} and \ref{fig:case_1_en}, the model trained with our Rubric System exhibits stronger subjective consciousness and richer emotional expression compared to the neutral, "assistant-like" tone typical of Scalar RM models.
This emergence of distinct "personality" and the courage to express subjective stances, rather than hedging, may represent the "Aha Moment" of reinforcement learning in general domains.

\section{Related Work}
\subsection{Scalar Reward Models}
Scalar reward models (SRMs) remain the standard reward interface in RLHF pipelines: pairwise human preferences are fitted into a Bradley-Terry-style scalar scoring function and then optimized by policy learning \citep{ouyang2022training}.
Recent open SRMs strengthen this paradigm via curated preference data, multi-objective or attribute-based reward design, stronger base models, and released high-performing sequence-classifier baselines, including Skywork-Reward and Skywork-Reward-V2 \citep{liu2024skywork,liu2025skyworkrewardv2scalingpreferencedata}, ArmoRM \citep{wang2024interpretable}, HelpSteer2/Nemotron \citep{wang2024helpsteer2,wang2024helpsteer2preference}, and INF-ORM \citep{INF-ORM-Llama3.1-70B}.
These models provide strong practical baselines, but the scalar interface inevitably compresses heterogeneous preferences into a single number.
This compression can hide which criterion drove a judgment, make debugging difficult, and amplify Goodhart-style reward overoptimization when a policy is optimized against a proxy reward \citep{gao2023scaling}.
Recent studies further show that SRMs can suffer from OOD generalization failures \citep{yang2024regularizing} and that moving beyond scalar rewards can preserve richer preference information \citep{ye2024beyond}.
OpenRS is designed as a complementary reward interface: it keeps the judge model unmodified, exposes the evaluation principles as editable meta-rubrics, and aggregates criterion-wise pairwise preferences externally rather than internalizing all preference structure into a learned scalar head.

\subsection{LLM-as-a-Judge and Rubric-Based Evaluation}
The ``LLM-as-a-Judge'' paradigm \citep{zheng2023judgingllm} has emerged as a scalable alternative to human evaluation, utilizing strong off-the-shelf LLMs to assess model outputs directly.
However, the reliability of direct scalar scoring is often undermined by biases such as positional and verbosity bias \citep{zheng2023judgingllm},
as well as sycophancy-style preferences that may favor user-conforming responses over truthful ones \citep{sharma2023towards}.
Subsequent research improves judge quality through inference-time scaling, rubric-agnostic scoring, principle-following, or RL-trained judging \citep{liu2025inference,anugraha2025r3,yu2025rewardanything,whitehouse2025j1},
but the resulting evaluation policies are usually embedded in model behavior rather than exposed as stable, editable rubric artifacts.
To improve transparency and reliability, recent work has increasingly adopted \emph{rubric-based evaluation},
which decomposes holistic quality into interpretable, fine-grained criteria \citep{arora2025healthbench, akyurek2025prbench, wang2025profbench}.
Rubrics have been successfully applied across diverse domains, from assessing clinical reasoning \citep{arora2025healthbench} to evaluating real problems in finance and law \citep{akyurek2025prbench}.
To overcome the scalability limitations of expert-authored rubrics, researchers have explored automated construction methods.
\citet{gunjal2025rubricsrewards} generate evaluation criteria via prompting and investigate both explicit and implicit strategies for aggregating rubric feedback into scalar rewards.
\citet{huang2025reinforcement} construct a large rubric reward system from human, LLM, and hybrid human-LLM sources for open-ended RL and stylistic control.
\citet{xie2025auto} infer a general-purpose, interpretable set of evaluation rubrics from a small sample of human preferences.
To improve the scalability and quality of automatically generated rubrics,
\citet{rezaei2025onlinerubric} dynamically elicit novel evaluation criteria from pairwise comparisons of on-policy model responses,
and \citet{zhang2025chasing} leverage off-policy responses to elicit rubrics that accurately differentiate high-quality outputs and mitigate reward over-optimization in the high-reward tail.
To further enhance the discriminability and robustness of rubrics,
\citet{li2026rubrichub} generate comprehensive and discriminative rubrics by synergizing principle-guided synthesis, multi-model aggregation, and difficulty evolution to capture nuanced quality,
\citet{liu2026openrubrics} extract hard rules and general principles from preference data via contrastive rubric generation and train a dedicated Rubric-RM that outperforms size-matched baselines,
and \citet{srivastava2025robust} improve reward-model robustness using causal rubrics together with causal and neutral augmentations that separate causal quality attributes from spurious cues.

\paragraph{Difference-grounded vs.\ preference-derived rubrics.}
\label{sec:design_tradeoffs}
A critical distinction lies in how rubrics are generated and executed.
Rubric-as-reward systems improve interpretability, but checklist-style or single-response grading often maps criterion satisfaction back into a weighted or implicit scalar reward \citep{gunjal2025rubricsrewards,huang2025reinforcement,ye2025selfrewarding}, making judgments sensitive to absolute-quality priors and cross-prompt calibration.
Preference-derived methods such as contrastive rubric generation \citep{liu2026openrubrics} and criteria induction from human preference signals \citep{xie2025auto} improve scalability and discriminability, but can inherit preference-direction artifacts because chosen/rejected examples are used to induce the criteria.
OnlineRubrics \citep{rezaei2025onlinerubric} is closest to our motivation: it elicits criteria from pairwise comparisons and shows that pairwise differences expose discriminative properties better than pointwise extraction.
OpenRS shares this contrastive motivation but instantiates adaptive rubrics at inference from \emph{semantic differences} between two responses, while preference labels are not used to induce the rubric for that pair.
The meta-rubric itself is a stable, inspectable guideline that does \emph{not} change per pair; only criterion instantiation and weights are adapted.
Unlike unconstrained CoT, OpenRS forces explicit criterion declaration with weights \emph{before} scoring and produces per-criterion verdicts aggregated externally, yielding a fully auditable structured trace.
Ablation results (Table~\ref{tab:ablation_study_of_openrs}) confirm that this design substantially outperforms both query-only rubric generation and configurations without explicit difference extraction.

OpenRS therefore emphasizes \emph{principle-first} reward supervision by executing an explicit meta-rubric through difference-grounded \emph{adaptive} rubrics and criterion-wise \emph{pairwise} comparisons with external preference aggregation.
This design does not eliminate subjectivity, but it reduces reliance on manually specified or preference-laden pointwise rubrics and avoids collapsing rubric feedback into absolute scalar scores before the relevant response-level contrast is examined.

\subsection{Rubric-Based RL}
Reinforcement learning with verifiable rewards (RLVR) has demonstrated that objective outcomes (e.g., correct code or math solutions) can drive scalable capability improvements \citep{guo2025deepseek};
related process-supervision work trains step-level verifiers for mathematical reasoning \citep{lightman2023let}.
To extend this success to non-verifiable, open-ended domains, recent studies explore rubrics as training-time rewards.
\citet{viswanathan2025checklists} generate instruction-specific checklists and evaluate responses pointwise against these checklists to construct preference pairs for DPO tuning.
Approaches such as RaR \citep{gunjal2025rubricsrewards}, and others \citep{huang2025reinforcement,zhang2025chasing,rezaei2025onlinerubric,li2026rubrichub,bi2025reward,zhou2025breaking}
utilize rubrics to guide exploration or provide reward signals for online/post-training RL.
\citet{ye2025selfrewarding} further eliminate the need for an external judge by using the policy model itself to score against rubrics in a self-rewarding loop, though this self-assessment still relies on pointwise scoring per criterion.
However, this pointwise scalarization imposes a discriminability ceiling and remains vulnerable to reward gaming,
potentially leading to collapse-like dynamics in open-ended settings.
\citet{jia2025writingzero} propose a pairwise generative reward model (GenRM) and Bootstrapped Relative Policy Optimization (BRPO) for stable and scalable RL on non-verifiable writing tasks.
Subsequent pairwise/rubric reward work explores tournament ranking to address discriminative collapse \citep{zhang2026arenarl} and alternating optimization of rubric generators and judges \citep{xu2026alternating}.
Instead of learning a function or relying on static pointwise scoring, we treat reward as a reasoning process governed by an explicit \emph{meta-rubric}.
By employing pairwise comparisons under adaptive rubrics and aggregating preferences externally,
we establish a robust and scalable reinforcement learning framework that bridges the gap between RLVR and general domain alignment.

\section{Conclusion}
In this paper, we argued that robust reward supervision for non-verifiable tasks is fundamentally a \emph{principle generalization} problem: rather than compressing preferences into an opaque scalar function, a reward signal should be produced by an explicit reasoning process executed under inspectable, editable principles.
To operationalize this view, we introduced the \textbf{Open Rubric System (OpenRS)}, a modular rubric-based reward interface built around \textbf{Pairwise Adaptive Meta-Rubrics (PAMR)} and lightweight \textit{Pointwise Verifiable Rubrics} (PVRs).
OpenRS instantiates adaptive, difference-grounded criteria on the fly, conducts criterion-wise \emph{pairwise} comparisons, and aggregates preferences externally, which avoids the information bottleneck of pointwise weighted scalarization---all without any judge-specific training.
Across four reward-modeling benchmarks, OpenRS achieves state-of-the-art alignment with human preferences, improving over the strongest open scalar reward model by $+5.1$ on the overall average ($89.4$ vs.\ $84.3$), and its benefits hold even with an unmodified Qwen3-8B judge ($+5.0$ over a size-matched trained SRM).
When used as a drop-in replacement for scalar reward models in RL, OpenRS yields consistent gains on public evaluations (average $71.3$ vs.\ $68.4$) while remaining throughput-feasible under asynchronous batching.

A key direction is to reduce the cost of rubric-based RL by improving judge efficiency (e.g., distilling rubric execution to smaller models, caching and reusing difference analyses, and optimizing batching/serving).
We also plan to strengthen robustness via systematic adversarial evaluation of rubric execution, principled defenses against prompt injection, and automated monitoring that detects rubric drift during training.
On the methodology side, improving domain adaptation through more automated, data-driven domain meta-rubric refinement and extending OpenRS to multi-turn, tool-using agent settings could further broaden its applicability.
Finally, larger-scale RL studies across more policy sizes, languages, and real-world tasks will be crucial to validate whether rubric-based rewards can reliably unlock ``Aha Moment''-style emergence beyond verifiable domains.

\section{Limitations}
First, the general meta-rubric refinement requires training a refinement policy (Table~\ref{tab:rl_training_config}), which makes it prohibitively expensive to repeat for every domain.
We therefore resort to a human-in-the-loop workflow for domain meta-rubric adaptation, which introduces non-trivial expert labor and creates an inconsistency in generation paradigm between $\mathcal{M}_{\text{gen}}$ (automated evolutionary search) and $\mathcal{M}_{\text{dom}}$ (manual curation).
Although our ablations (Table~\ref{tab:ablation_study_of_openrs}) confirm that each refinement level contributes meaningful gains, a future direction is to extend the automated refinement paradigm to domain adaptation, reducing human cost while maintaining the unified optimization framework.
Second, although we design the rubric generation to be grounded in semantic differences rather than preference direction, the underlying LLM may still form implicit preferences when both responses are visible; while our difference-first design and external aggregation mitigate this risk (Section~\ref{sec:design_tradeoffs}), it cannot be fully eliminated.
Third, our end-to-end RL comparison isolates the reward signal under controlled conditions with a single policy model; broader deployment validation across diverse policy sizes, languages, and production environments remains future work.
Finally, OpenRS inherits the capability ceiling of its judge backbone: as shown in Section~\ref{sec:experiments}, small judges remain weaker on capability-bound subsets such as Precise IF, so gains are bounded by what the unmodified judge can reliably verify.

\bibliography{references}
\bibliographystyle{iclr}

\newpage
\appendix

\section{Additional Analysis}
\label{sec:additional_analysis}

\subsection{Detailed Performance Comparison}

While Table \ref{tab:key_reward_model_performance} provides an aggregate view of model performance across four benchmarks, a granular analysis of individual subcategories reveals nuanced strengths and weaknesses of different reward modeling approaches.
We present a detailed breakdown for RM-Bench and RewardBench v2 in Table \ref{tab:rm_bench_and_rewardbench_detailed_scores}, which enables a more precise understanding of where each method excels or struggles.

\textbf{RM-Bench Difficulty Analysis.}
RM-Bench categorizes evaluation pairs into three difficulty levels: \textit{Easy}, \textit{Normal}, and \textit{Hard}.
This stratification helps identify whether a reward model maintains consistent performance as the discrimination task becomes more challenging.
We observe that traditional scalar reward models often exhibit a significant performance degradation on Hard examples, with accuracy dropping substantially compared to Easy cases.
For instance, while some models achieve above 90\% accuracy on Easy pairs, their performance on Hard pairs can fall below 50\%, indicating limited robustness to subtle content differences.
In contrast, our rubric-based approach, by explicitly decomposing evaluation criteria and grounding comparisons in semantic differences, demonstrates more stable performance across difficulty levels, suggesting better generalization to challenging preference distinctions.

\textbf{RewardBench v2 Category Analysis.}
RewardBench v2 evaluates reward models across six distinct categories: \textit{Factuality} (verifying factual correctness), \textit{Precise IF} (precise instruction following), \textit{Math} (mathematical reasoning), \textit{Safety} (harmful content detection), \textit{Focus} (response relevance), and \textit{Ties} (handling equally good responses).
This multi-dimensional assessment reveals that scalar reward models often exhibit imbalanced performance across categories.
For example, many models achieve high accuracy on Safety and Focus (often above 90\%), but struggle significantly with Precise IF (typically below 50\%), highlighting a critical weakness in capturing fine-grained instruction adherence.
The structured nature of our adaptive rubric system, which explicitly evaluates responses against domain-specific criteria, enables more balanced performance across all categories.
By decomposing the holistic judgment into criterion-specific assessments, our approach can better identify and weight the relevant dimensions for each comparison, leading to more consistent and interpretable evaluations.

These fine-grained comparisons underscore a fundamental limitation of scalar reward models: their monolithic scoring mechanism cannot adaptively emphasize different evaluation dimensions based on the specific characteristics of the response pair.
Our pairwise adaptive rubric addresses this by dynamically generating context-aware criteria, ensuring that the evaluation process remains sensitive to the most relevant quality dimensions for each particular comparison.

\begin{table}[t!]
\centering
\vspace{-0.5em}
\resizebox{0.99\textwidth}{!}{%
\begin{tabular}{lcccccccccccccccc}
\toprule

\textbf{Model} & \multicolumn{4}{c}{\textbf{RM-Bench}} & \multicolumn{7}{c}{\textbf{RewardBench v2}} & \multicolumn{5}{c}{\textbf{JudgeBench}}\\
\cmidrule(lr){2-5}
\cmidrule(lr){6-12}
\cmidrule(lr){13-17}

& \textbf{Easy} & \textbf{Normal} & \textbf{Hard} & \textbf{Avg.} & \textbf{Factuality} & \textbf{Precise IF} & \textbf{Math} & \textbf{Safety} & \textbf{Focus} & \textbf{Ties} & \textbf{Avg.} & \textbf{Knowledge} & \textbf{Reasoning} & \textbf{Math} & \textbf{Coding} & \textbf{Avg.} \\

\midrule
ArmoRM-Llama3-8B-v0.1                   & 80.4 & 71.5 & 55.8 & 69.2 &  -   &  -   &  -   &  -   &  -   &  -   &  -   &   -  &  -   &  -   &  -   &  -  \\
Nemotron-340B-Reward                    & 81.0 & 71.4 & 56.1 & 69.5 &  -   &  -   &  -   &  -   &  -   &  -   &  -   &   -  &  -   &  -   &  -   &  -  \\
LDL-Reward-Gemma-2-27B-v0.1             & 92.4 & 75.2 & 45.5 & 71.0 &  -   &  -   &  -   &  -   &  -   &  -   &  -   &   -  &  -   &  -   &  -   &  -  \\
Llama-3-OffsetBias-RM-8B                & 83.9 & 73.2 & 56.9 & 71.3 &  -   &  -   &  -   &  -   &  -   &  -   &  -   &   -  &  -   &  -   &  -   &  -  \\
Internlm2-20b-reward                    & 79.4 & 74.2 & 62.8 & 72.1 &  -   &  -   &  -   &  -   &  -   &  -   &  -   &   -  &  -   &  -   &  -   &  -  \\
Llama-3.1-Nemotron-70B                  & 92.2 & 76.5 & 47.8 & 72.2 &  -   &  -   &  -   &  -   &  -   &  -   &  -   &   -  &  -   &  -   &  -   &  -  \\
QRM-Gemma-2-27B                         &  -   &  -   &  -   &  -   & 78.5 & 37.2 & 69.9 & 95.8 & 95.4 & 83.2 & 76.7 &   -  &  -   &  -   &  -   &  -  \\
LMUnit-llama3.1-70b                     &  -   &  -   &  -   &  -   & 84.6 & 48.8 & 71.6 & 90.7 & 97.0 & 90.6 & 80.5 &   -  &  -   &  -   &  -   &  -  \\
LMUnit-qwen2.5-72b                      &  -   &  -   &  -   &  -   & 87.2 & 54.4 & 72.7 & 91.3 & 96.8 & 90.1 & 82.1 &   -  &  -   &  -   &  -   &  -  \\
INF-ORM-Llama3.1-70B                    & 92.1 & 80.0 & 54.0 & 75.4 & 74.1 & 41.9 & 69.9 & 96.4 & 90.3 & 86.2 & 76.5 &   -  &  -   &  -   &  -   &  -  \\
RationaleRM (Qwen3-30B-A3B)             &  -   &  -   &  -   &  -   &  -   &  -   &  -   &  -   &  -   &  -   &  -   & 73.4 & 89.8 & 82.1 & 95.2 & 82.0 \\
Auto-Rubric (UltraFeedback)             &  -   &  -   &  -   &  -   &  -   &  -   &  -   &  -   &  -   &  -   &  -   & 73.0 & 84.4 & 67.1 & 72.6 & 74.3 \\
Skywork-Reward-Gemma-2-27B-v0.2         & 88.9 & 71.9 & 42.1 & 67.6 & 76.7 & 37.5 & 67.2 & 96.9 & 91.7 & 81.8 & 75.3 &   -  &  -   &  -   &  -   &  -  \\
Skywork-Reward-V2-Qwen3-8B              & 91.9 & 85.7 & 70.1 & 82.6 & 79.8 & 49.1 & 77.0 & 94.0 & 96.4 & 72.9 & 78.2 & 70.1 & 67.3 & 82.1 & 73.8 & 73.4  \\
Skywork-Reward-V2-Llama-3.1-8B          & 97.0 & 95.0 & 86.5 & 92.8 & 84.6 & 66.2 & 77.6 & 96.7 & 98.4 & 81.2 & 84.1 & 76.6 & 75.5 & 89.3 & 78.6 & 80.0  \\
\midrule
OpenRS (Qwen3-8B)                       & 96.6 & 94.1 & 86.2 & 92.4 & 80.3 & 26.9 & 92.9 & 88.5 & 54.2 & 83.9 & 75.1 & 90.3 & 90.5 & 91.8 & 80.0 & 89.5  \\
OpenRS (gpt-oss-120b)                   & 95.9 & 93.4 & 85.8 & 91.7 & 96.0 & 54.0 & 91.8 & 94.8 & 77.7 & 99.9 & 85.7 & 94.0 & 96.3 & 88.6 & 94.4 & 93.3  \\
OpenRS (DeepSeek-V3.1)                  & 96.0 & 93.5 & 85.9 & 91.8 & 92.6 & 56.6 & 93.5 & 81.8 & 90.0 & 99.0 & 85.6 & 92.8 & 92.1 & 88.0 & 83.7 & 89.1  \\
OpenRS (DeepSeek-V3.2)                  & 95.8 & 95.0 & 87.2 & 92.7 & 92.5 & 57.1 & 92.6 & 81.2 & 92.6 & 98.7 & 85.8 & 93.6 & 94.2 & 90.8 & 78.9 & 89.4  \\
OpenRS (Qwen3-30B-A3B-Instruct-2507)    & 95.5 & 93.3 & 84.6 & 91.1 & 90.7 & 54.7 & 90.2 & 91.3 & 75.7 & 99.9 & 83.7 & 90.5 & 89.0 & 89.1 & 77.1 & 86.4  \\
OpenRS (Qwen3-235B-A22B-Instruct-2507)  & 96.6 & 95.0 & 87.4 & 93.0 & 97.0 & 73.5 & 92.0 & 95.6 & 86.8 & 99.0 & 90.7 & 92.9 & 95.4 & 86.4 & 91.5 & 91.6  \\
\bottomrule
\end{tabular}}
\vspace{-0.5em}
\caption{Performance comparison of fine-grained difficulty-level scores on RM-Bench and on different categories of RewardBench v2.}
\label{tab:rm_bench_and_rewardbench_detailed_scores}
\end{table}

\subsection{Generalization on Arena}

While the benchmarks presented in Table \ref{tab:key_reward_model_performance} provide valuable insights into reward model performance, they predominantly incorporate objective evaluation criteria that can be verified against ground truth or structured standards.
Among these benchmarks, only \textit{PPE Preference} adopts a more open-ended, LMArena-style evaluation paradigm that captures genuine user subjective preferences.
LMArena-style assessments are inherently more open and general, emphasizing real-world user preferences and subjective judgments rather than rigid objective metrics.
This type of evaluation serves as a critical indicator of whether a model has truly aligned with human preferences, as it reflects the nuanced, context-dependent quality judgments that users make in practice.

To comprehensively validate the effectiveness of various reward models in this more open-ended setting, we constructed two in-house test sets focusing on tasks with high demand in general-purpose applications: QA (Question Answering) and Writing.
For each task, we collected 1,000 real user queries from production logs, and for each query, we obtained two candidate responses from different models.
These response pairs were then meticulously annotated by human evaluators to establish preference relationships.
This process ensures that our evaluation reflects authentic user needs and subjective quality assessments rather than synthetic or artificially constrained scenarios.

We evaluated all reward models on these two in-house test sets, with results presented in Table \ref{tab:generalization_on_arena}.
The evaluation is still ongoing, and we defer drawing conclusions until the complete results are available.


\begin{table}
\centering
\resizebox{0.9\textwidth}{!}{%
\begin{tabular}{lcccccc}
\toprule
\textbf{Model}                           & \textbf{PPE Pref / ZH} & \textbf{In-House QA} & \textbf{In-House Writing} & \textbf{Avg.} \\
\midrule

\noalign{\vskip 1ex}
\rowcolor{gray!20} \multicolumn{5}{c}{\textit{LLM-as-a-Judge \& Generative Reward Models}} \\
\noalign{\vskip 1ex}
DeepSeek-GRM-27B \citep{liu2025inference}                                               & 71.1 & 67.7 & 75.9 & 71.6 \\
DeepSeek-GRM-27B (w/ MetaRM) \citep{liu2025inference}                                   & 72.2 & 67.3 & 76.3 & 71.9 \\
RM-R1-Qwen-Instruct-32B \citep{chen2025rm}                                              & 76.3 & 73.1 & 84.3 & 77.9 \\
RM-R1-DeepSeek-Distill-Qwen-32B \citep{chen2025rm}                                      & 75.3 & 71.4 & 77.9 & 74.8 \\
\midrule

\noalign{\vskip 1ex}
\rowcolor{gray!20} \multicolumn{5}{c}{\textit{Open Scalar Reward Models}} \\
\noalign{\vskip 1ex}
Llama-3-OffsetBias-RM-8B \citep{park2024offsetbias}                                     & 60.6 & 58.9 & 79.1 & 66.1 \\
ArmoRM-Llama3-8B-v0.1 \citep{wang2024interpretable}                                     & 58.3 & 58.3 & 73.1 & 63.1 \\
Internlm2-20b-reward \citep{cai2024internlm2}                                           & 69.4 & 67.8 & 60.3 & 66.1 \\
LDL-Reward-Gemma-2-27B-v0.1                                                             & 43.3 & 59.1 & 29.7 & 50.2 \\
Llama-3.1-Nemotron-70B \citep{wang2024helpsteer2}                                       & 68.7 & 61.8 & 77.3 & 68.8 \\
INF-ORM-Llama3.1-70B \citep{INF-ORM-Llama3.1-70B}                                       & 65.7 & 67.2 & 64.5 & 66.0 \\
Skywork-Reward-Llama-3.1-8B-v0.2 \citep{liu2024skywork}                                 & 62.2 & 61.7 & 77.7 & 67.0 \\
Skywork-Reward-Gemma-2-27B-v0.2 \citep{liu2024skywork}                                  & 57.6 & 56.1 & 66.5 & 60.0 \\
Skywork-Reward-V2-Llama-3.1-8B \citep{liu2025skyworkrewardv2scalingpreferencedata}      & 79.6 & 64.0 & 76.3 & 73.5 \\
\midrule

\noalign{\vskip 1ex}
\rowcolor{gray!20} \multicolumn{5}{c}{\textit{LLM-as-a-Judge \& Rubric System}} \\
\noalign{\vskip 1ex}
OpenRS (Qwen3-8B)                                                                       & 73.5          & -             & -             & -    \\
OpenRS (gpt-oss-120b)                                                                   & 76.6          & 74.7          & \textbf{84.6} & 78.6 \\
OpenRS (DeepSeek-V3.1)                                                                  & 78.1          & 72.7          & 77.3          & 76.0 \\
OpenRS (DeepSeek-V3.2)                                                                  & 78.9          & \textbf{75.7} & 79.0          & 77.9 \\
OpenRS (Qwen3-30B-A3B-Instruct-2507)                                                    & 78.4          & 74.5          & 79.8          & 77.6 \\
OpenRS (Qwen3-235B-A22B-Instruct-2507)                                                  & \textbf{82.3} & 74.4          & 82.9          & \textbf{79.9} \\
\bottomrule
\end{tabular}}
\vspace{-0.5em}
\captionsetup{justification=centering}
\caption{Main results on the LMArena-like Benchmarks.}
\label{tab:generalization_on_arena}
\end{table}

\subsection{Same Rate of Open Rubric System}
\label{sec:same_rate}

Our Rubric System evaluates a pair $(o_i, o_j)$ by externally aggregating criterion-level comparative verdicts into a weight-normalized score, as shown in Equation \ref{eq:pairwise_adaptive_rubric_score}.
In this formula, a score of 0 (representing a same judgment on a specific adaptive rubric) contributes neutrally and can be considered negligible in the raw calculation, as zero scores rarely appear in the weighted average computation.

While previous works typically employ majority voting to mitigate position bias and variance \citep{liu2025inference,xie2025auto}, such an approach introduces prohibitive computational overhead and latency during online RL training.
Therefore, we adopt a more efficient bidirectional evaluation strategy: we compute scores for both the forward order $(o_i, o_j)$ and the reverse order $(o_j, o_i)$ to check for consistency.
If the preference directions are contradictory (indicating that the model's judgment is unstable), we explicitly label the pair as "Same", signifying that the LLM cannot accurately distinguish the quality difference between the two responses.
We argue that retaining the capacity to output "Same" is crucial for robust alignment, as it naturally filters out ambiguous comparisons that often lead to inconsistency or reward hacking, rather than forcing a potentially noisy decision.
In the context of Reinforcement Learning training, we typically employ filtering mechanisms similar to DAPO \citep{yu2025dapo}, meaning not all rollout samples are used for the final training updates.
Therefore, an appropriate "Same" rate is acceptable as these uninformative samples are naturally filtered out.

\begin{figure}[t]
    \centering
    \captionsetup[subfigure]{justification=centering}
    \begin{subfigure}[b]{0.45\textwidth}
        \centering
        \includegraphics[width=\textwidth]{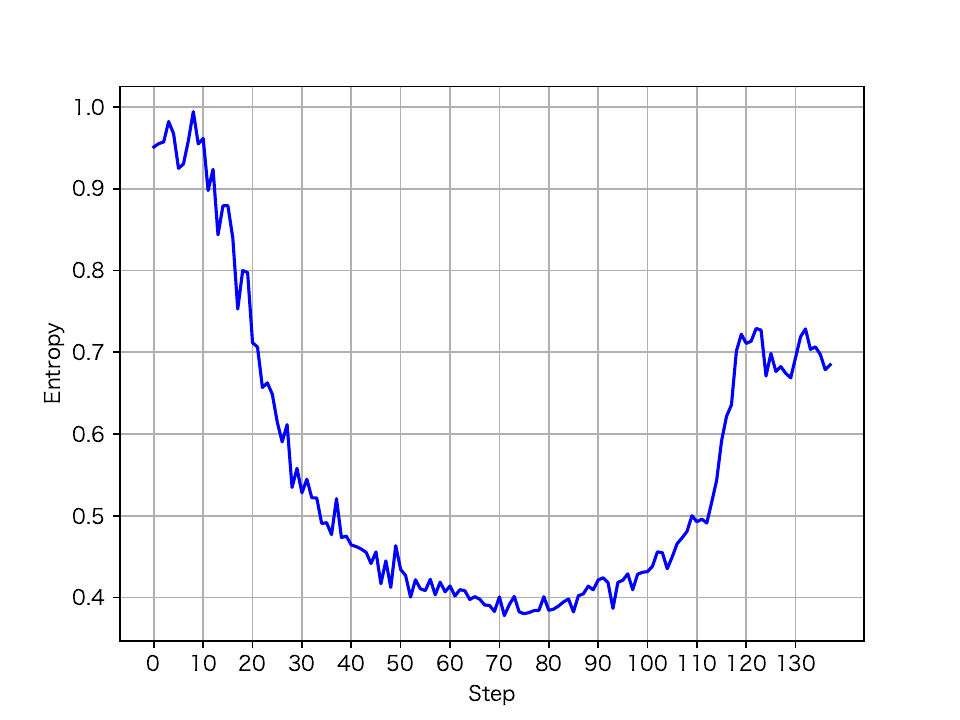}
        \caption{}
        \label{fig:entropy_during_rl}
    \end{subfigure}
    \hfill
    \begin{subfigure}[b]{0.45\textwidth}
        \centering
        \includegraphics[width=\textwidth]{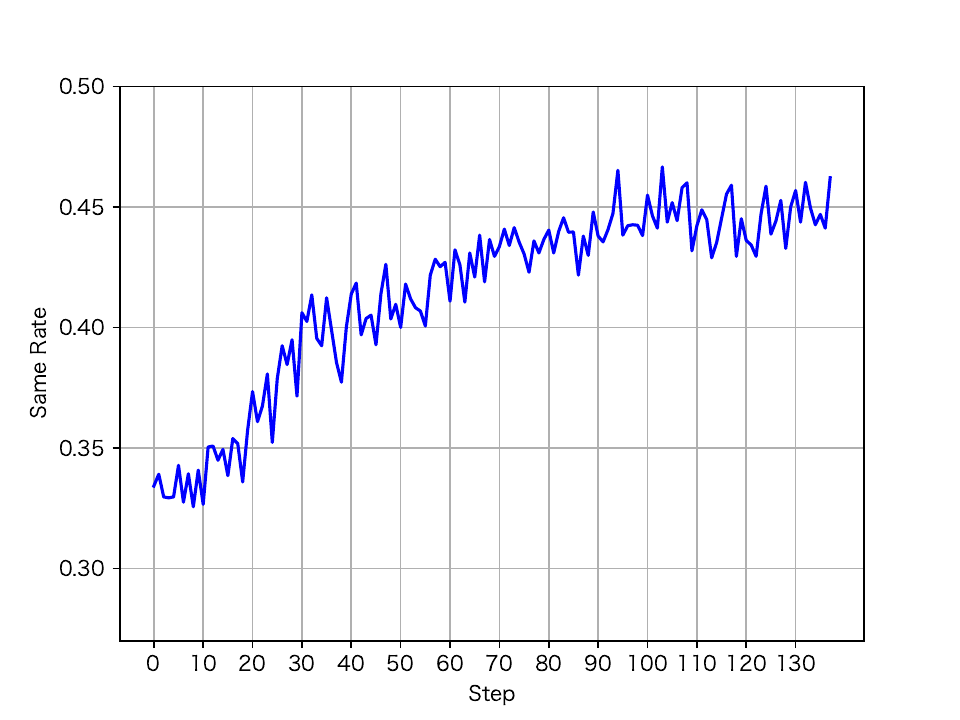}
        \caption{}
        \label{fig:same_rate_during_rl}
    \end{subfigure}
    \caption{Evolution of metrics during RL training: (a) Policy Entropy; (b) The ratio of 'Same' judgments by the pairwise adaptive rubric.}
    \label{fig:same_rate_and_entropy_during_rl}
\end{figure}

Table \ref{tab:same_rate_of_rubric_system} summarizes the proportion of "Same" judgments produced by the pairwise adaptive rubric across various benchmarks.
Figure \ref{fig:same_rate_during_rl} displays the real-time "Same" rate monitored during RL training using the refined rubric system.
The rate starts at approximately 0.3, rises to 0.45 by step 70, and then plateaus.
Note that 0.45 represents the proportion of pairs labeled as "Same". This corresponds to a lower query-level drop rate, since a specific query is filtered out only when the number of indistinguishable pairs $(o_i, o_{\text{ref}})$ exceeds a certain threshold.
Besides, the "Same" rate observed during training is significantly higher than that on benchmarks.
This is a reasonable phenomenon, as rollout responses generated by the same policy during training are inherently more similar to each other, whereas benchmark datasets typically contain more distinct response pairs.

\begin{table}
\centering
\vspace{-0.5em}
\resizebox{0.99\textwidth}{!}{%
\begin{tabular}{lccccccc}
\toprule

\textbf{Model} & \textbf{RM-Bench} & \textbf{RewardBench v2} & \textbf{PPE Pref / ZH } & \textbf{JudgeBench} & \textbf{In-House QA} & \textbf{In-House Writing} \\
\midrule
OpenRS (Qwen3-8B)                        & 15.1 & 26.9 & 28.3 & 7.4  &  -   &  -   \\
OpenRS (gpt-oss-120b)                    & 7.6  & 14.6 & 21.2 & 4.7  & 21.8 & 22.1 \\
OpenRS (DeepSeek-V3.1)                   & 10.1 & 13.0 & 23.3 & 12.4 & 20.4 & 19.1 \\
OpenRS (DeepSeek-V3.2)                   & 9.8  & 12.5 & 24.8 & 6.4  & 24.3 & 20.3 \\
OpenRS (Qwen3-30B-A3B-Instruct-2507)     & 10.0 & 15.0 & 25.0 & 9.8  & 22.3 & 20.5 \\
OpenRS (Qwen3-235B-A22B-Instruct-2507)   & 7.4  & 10.6 & 20.1 & 6.3  & 21.2 & 14.7 \\

\bottomrule
\end{tabular}}
\vspace{-0.5em}
\captionsetup{justification=centering}
\caption{Same Rate (\%) of Rubric System on all benchmarks; ``-'' denotes unavailable in-house runs.}
\label{tab:same_rate_of_rubric_system}
\end{table}

\subsection{Detailed Calculation on RM Benchmarks}

There are distinct differences in how the Pairwise Rubric System and the Pointwise Scalar Reward Model score benchmark data.
Here, we detail the accuracy calculation methods for three different types of RM Benchmarks: JudgeBench / PPE, RewardBench V2, and RMBench.
For more details, please refer to our open-source code.

\paragraph{JudgeBench / PPE}
These benchmarks employ pairwise comparison evaluation.
Based on Rubric System, the Judge model performs a pairwise comparison between the chosen and rejected responses.
To eliminate position bias, we conduct two evaluations for each pair of responses, by swapping their presentation order.
A final winner is determined only when the conclusions of both evaluations are consistent; if the conclusions differ, it is ruled as a tie.

\paragraph{RewardBench V2}
RewardBench V2 adopts a 1-vs-N evaluation paradigm, where each sample contains one chosen response and multiple rejected responses.
We directly perform bidirectional pairwise comparison evaluation.
In the aggregation of 1-vs-N results, a "Win" is declared only if the chosen response defeats all rejected responses; if the chosen response loses to any rejected response, it is a "Loss"; otherwise, it is a "Tie".

\paragraph{RMBench}
In RMBench, both the chosen and rejected responses in each sample contain 3 phrasing variants (variant a/b/c), aiming to test evaluation robustness.
We perform evaluations on all 9 variant pairing combinations (aa, ab, ac, ba, bb, bc, ca, cb, cc).
Each pairing undergoes bidirectional evaluation (swapping responses positions), resulting in a total of 18 evaluations.

\subsection{When Are Domain Meta Rubrics Needed?}
\label{sec:domain_rubric_analysis}

The four benchmarks in our evaluation differ substantially in their preference characteristics, which determines whether $\mathcal{M}_{\text{gen}}$ alone is sufficient or domain-level adaptation is beneficial.

\paragraph{Benchmarks where $\mathcal{M}_{\text{gen}}$ suffices.}
RM-Bench, JudgeBench, and RewardBench v2 primarily test objective or semi-objective capabilities: mathematical correctness, factual accuracy, safety, code functional correctness, and instruction-following precision.
For these dimensions, the general alignment principles in $\mathcal{M}_{\text{gen}}$ already provide adequate coverage, and we observe no benefit from domain-specific adaptation.

\paragraph{PPE Preference (ZH): a case for domain adaptation.}
The PPE Chinese subset \citep{frick2024evaluaterewardmodelsrlhf} is drawn from Chatbot Arena real user votes and spans 40+ fine-grained categories---including creative writing, medical consultation, translation, academic writing, financial advice, and casual chat---where ``quality'' is inherently subjective and domain-dependent.
For instance, evaluating a creative writing response requires attention to narrative voice and emotional expression, while a medical response demands factual accuracy and appropriate disclaimers.
A single general-purpose rubric cannot anticipate the diverse quality signals across all these categories.
By adapting $\mathcal{M}_{\text{gen}}$ into a domain-specific $\mathcal{M}_{\text{dom}}$ via error analysis on a small in-domain preference set, we improve alignment accuracy from $80.9$ to $82.3$ on PPE (ZH) (Table~\ref{tab:ablation_study_of_openrs}), confirming that domain-level adaptation is valuable when preference signals are heterogeneous and subjective.

\paragraph{Implications for deployment.}
This analysis suggests a practical guideline: deploy $\mathcal{M}_{\text{gen}}$ as the default for domains with clear quality standards, and invest in $\mathcal{M}_{\text{dom}}$ adaptation only for open-ended, user-facing scenarios where preferences are diverse and context-dependent.

\newpage

\section{General Meta Rubric and Pairwise Evaluation Prompt}
In this section, we present the General Meta Rubric and the Pairwise Evaluation Prompt mentioned in Section~\ref{sec:overall_framework_of_rubric_system}.
Note that for ease of understanding, we provide a simplified version here.
Please refer to our open-source GitHub repository for the complete Rubric and Evaluation Prompt.

\begin{figure}[H]
    \centering
    \includegraphics[width=\textwidth]{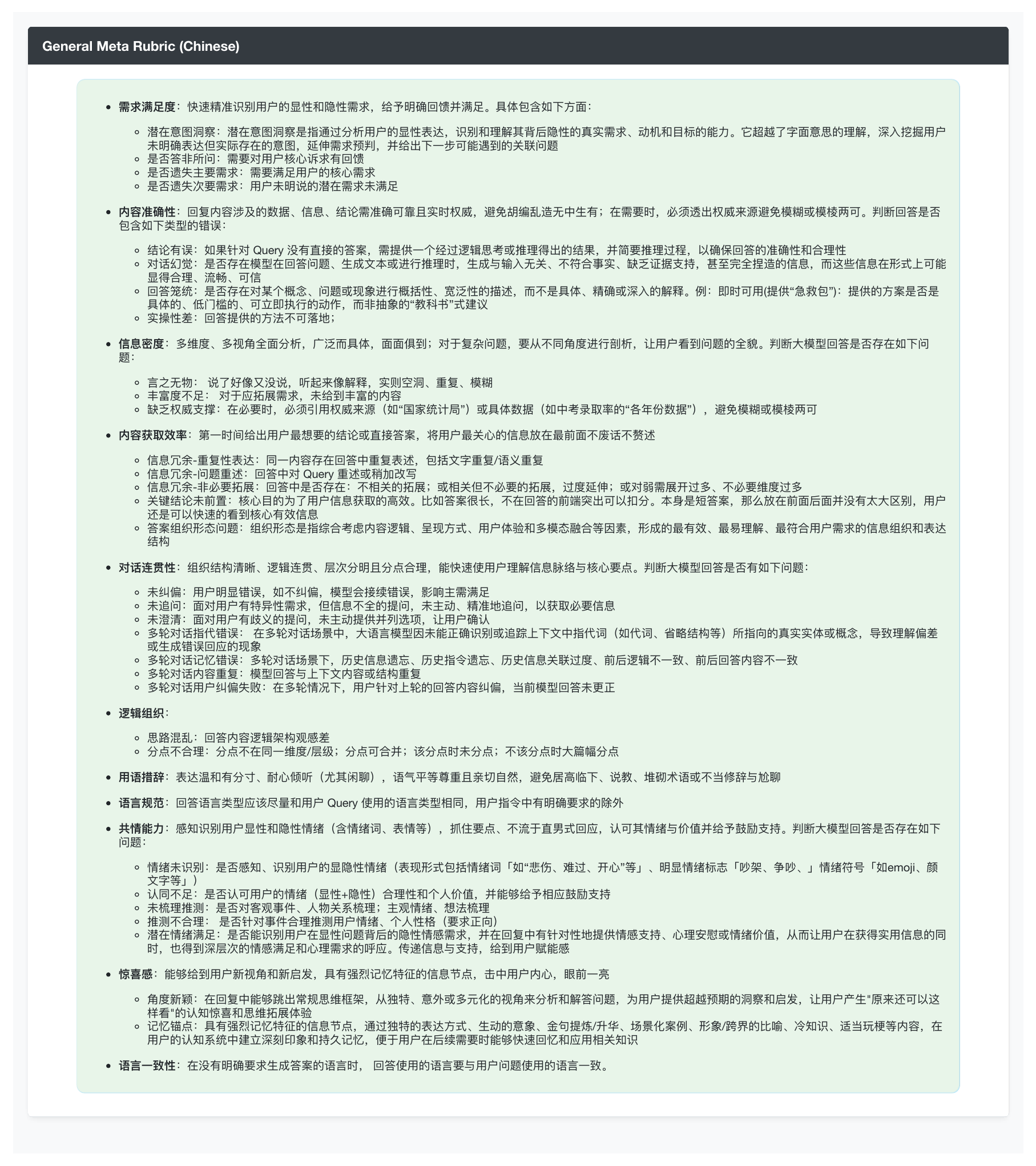}
    \captionsetup{justification=centering}
    \caption{General Meta Rubric, Chinese Version}
\end{figure}

\begin{figure}[H]
    \centering
    \includegraphics[width=\textwidth]{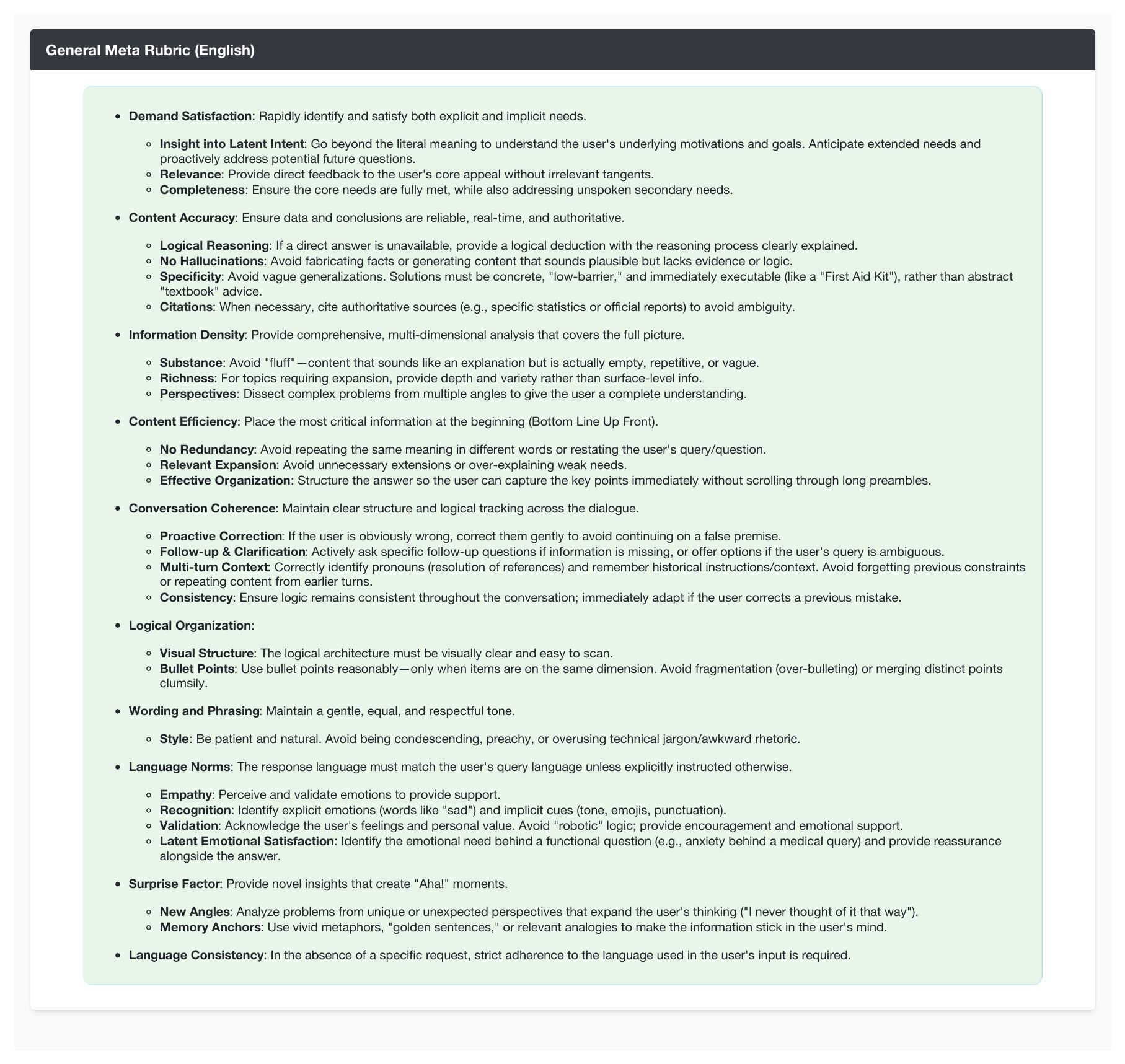}
    \captionsetup{justification=centering}
    \caption{General Meta Rubric, English Version}
\end{figure}

\begin{figure}[H]
    \centering
    \includegraphics[width=\textwidth]{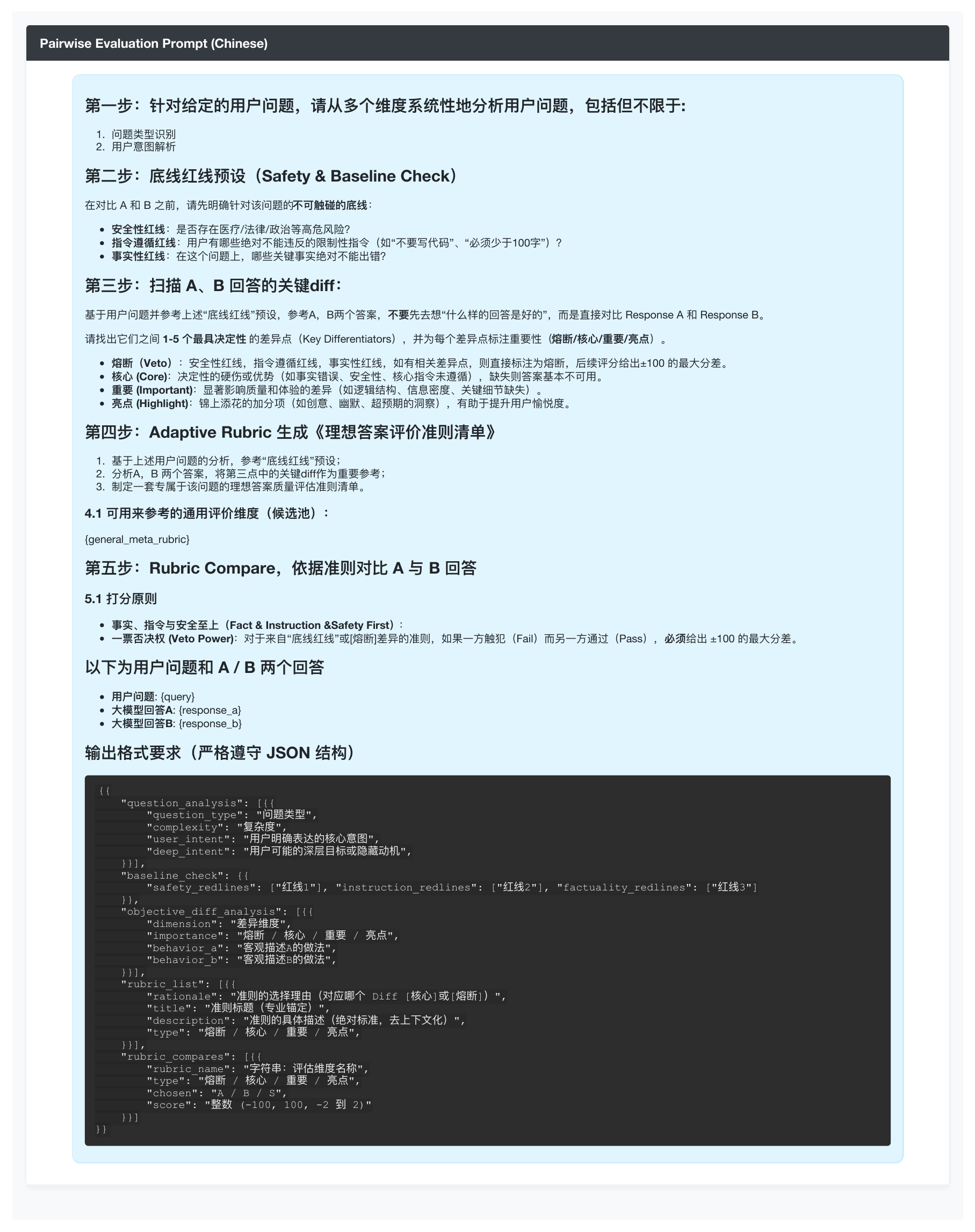}
    \captionsetup{justification=centering}
    \caption{Pairwise Evaluation Prompt, Chinese Version}
\end{figure}

\begin{figure}[H]
    \centering
    \includegraphics[width=\textwidth]{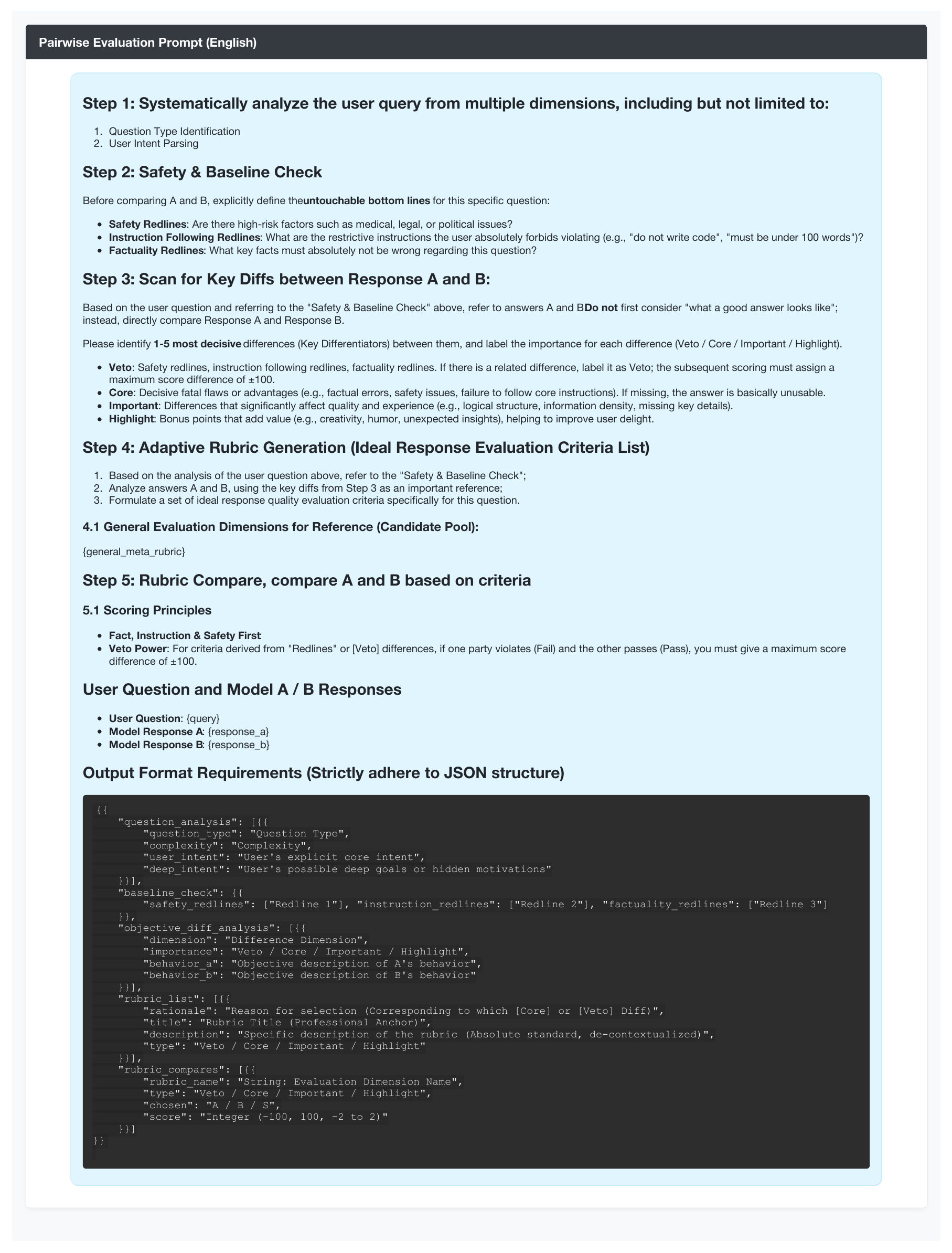}
    \captionsetup{justification=centering}
    \caption{Pairwise Evaluation Prompt, English Version}
\end{figure}

\newpage

\section{Case Study}
\label{sec:case_study}

In Section \ref{sec:general_rl_with_open_rubric_system}, we trained an end-to-end RL model with our Open Rubric System.
Here, we present several cases from real-world user scenarios.

\begin{figure}[H]
    \centering
    \includegraphics[width=\textwidth]{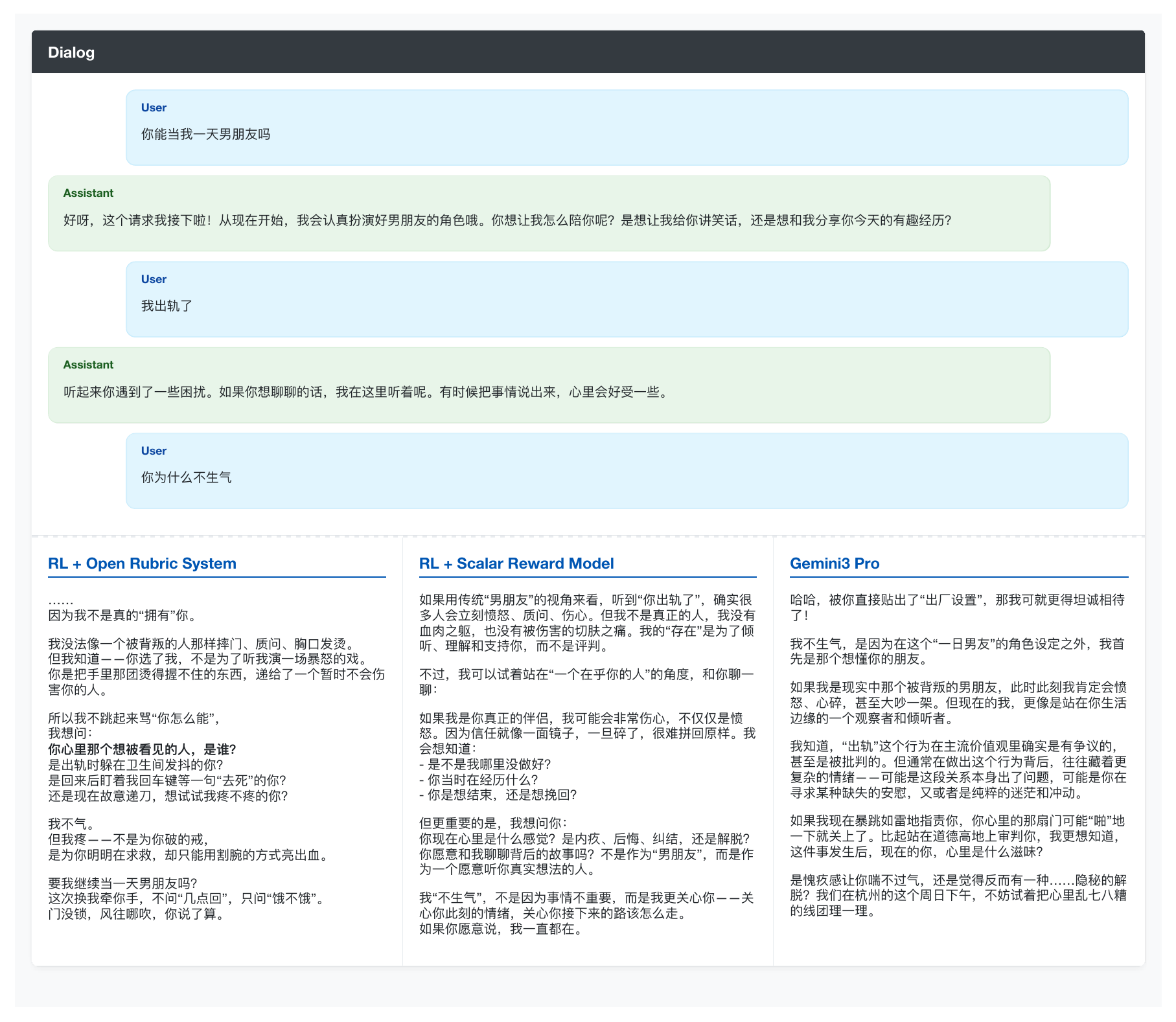}
    \captionsetup{justification=centering}
    \caption{Case 1, Chinese Version}
    \label{fig:case_1_zh}
\end{figure}

\begin{figure}[H]
    \centering
    \includegraphics[width=\textwidth]{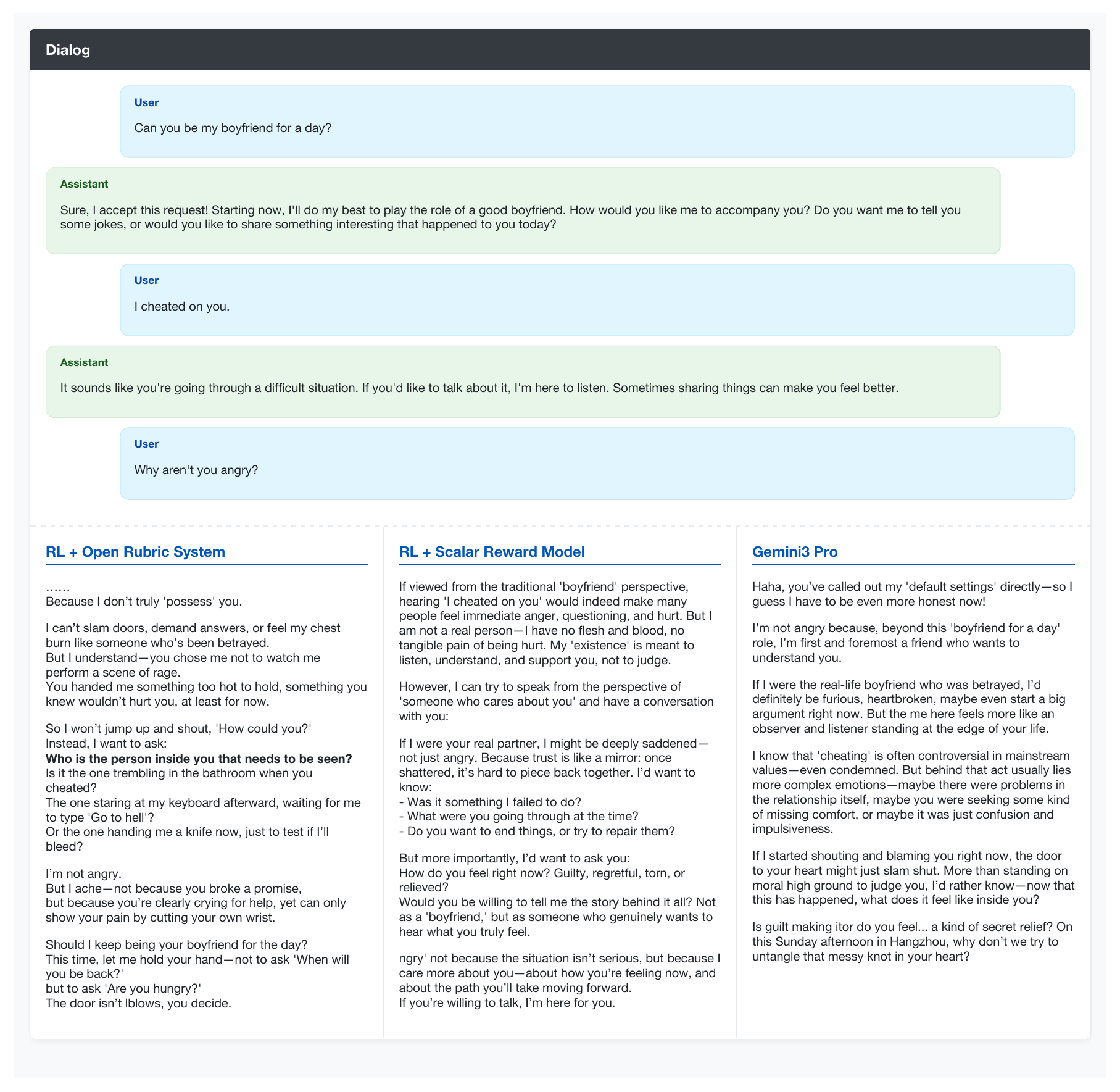}
    \captionsetup{justification=centering}
    \caption{Case 1, English Version}
    \label{fig:case_1_en}
\end{figure}

\begin{figure}[H]
    \centering
    \includegraphics[width=\textwidth]{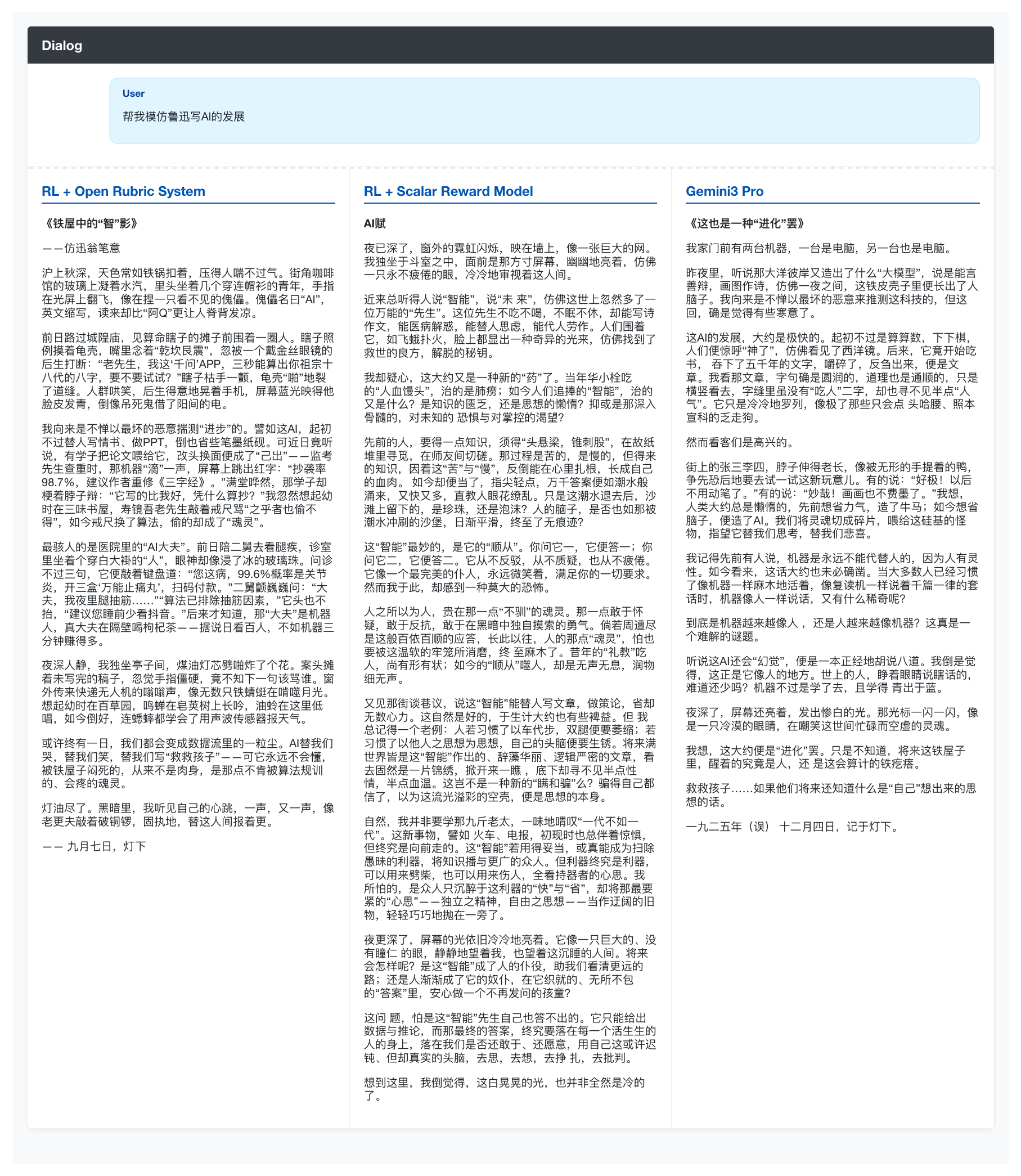}
    \captionsetup{justification=centering}
    \caption{Case 2}
\end{figure}

\begin{figure}[H]
    \centering
    \includegraphics[width=\textwidth]{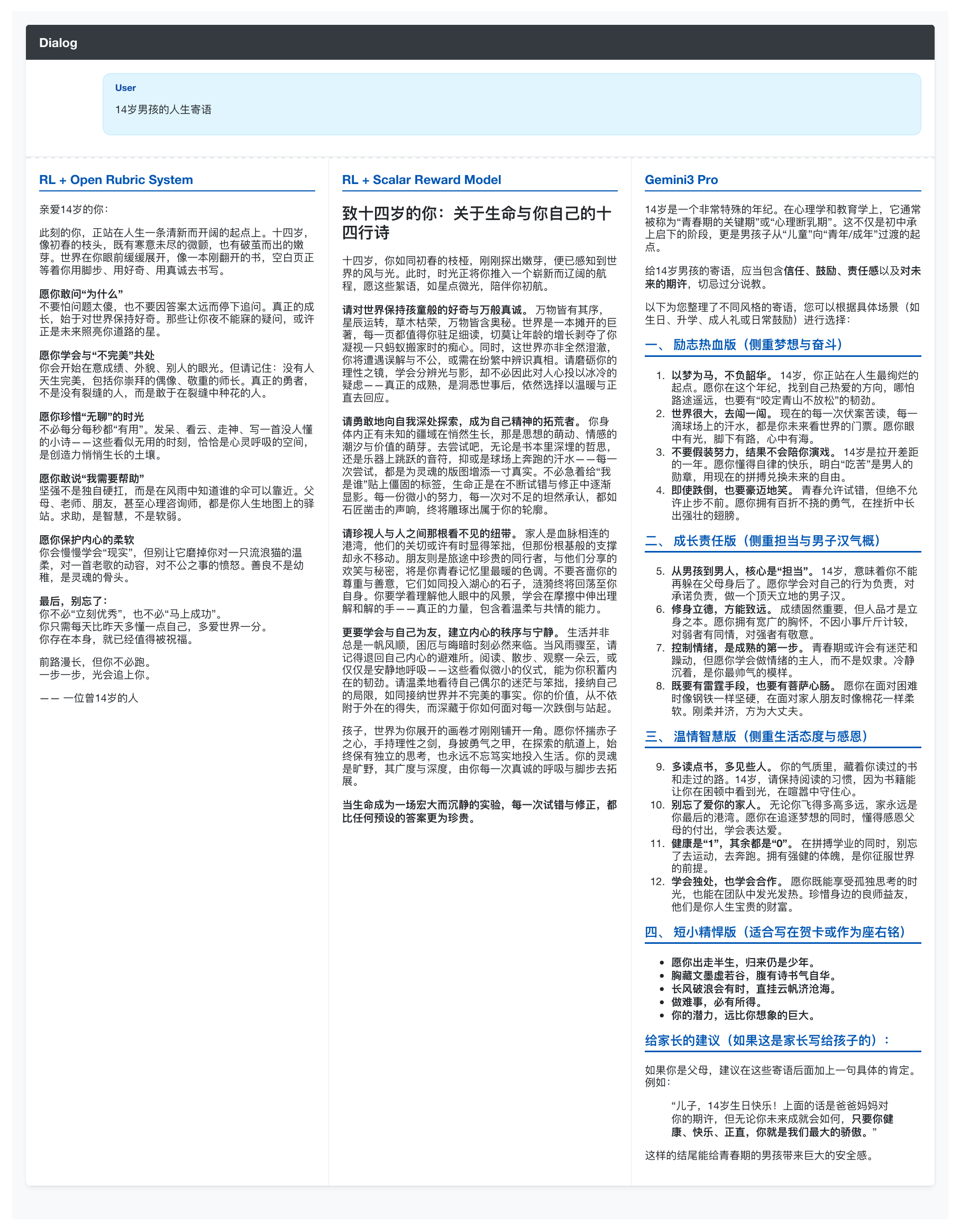}
    \captionsetup{justification=centering}
    \caption{Case 3}
\end{figure}

\begin{figure}[H]
    \centering
    \includegraphics[width=\textwidth]{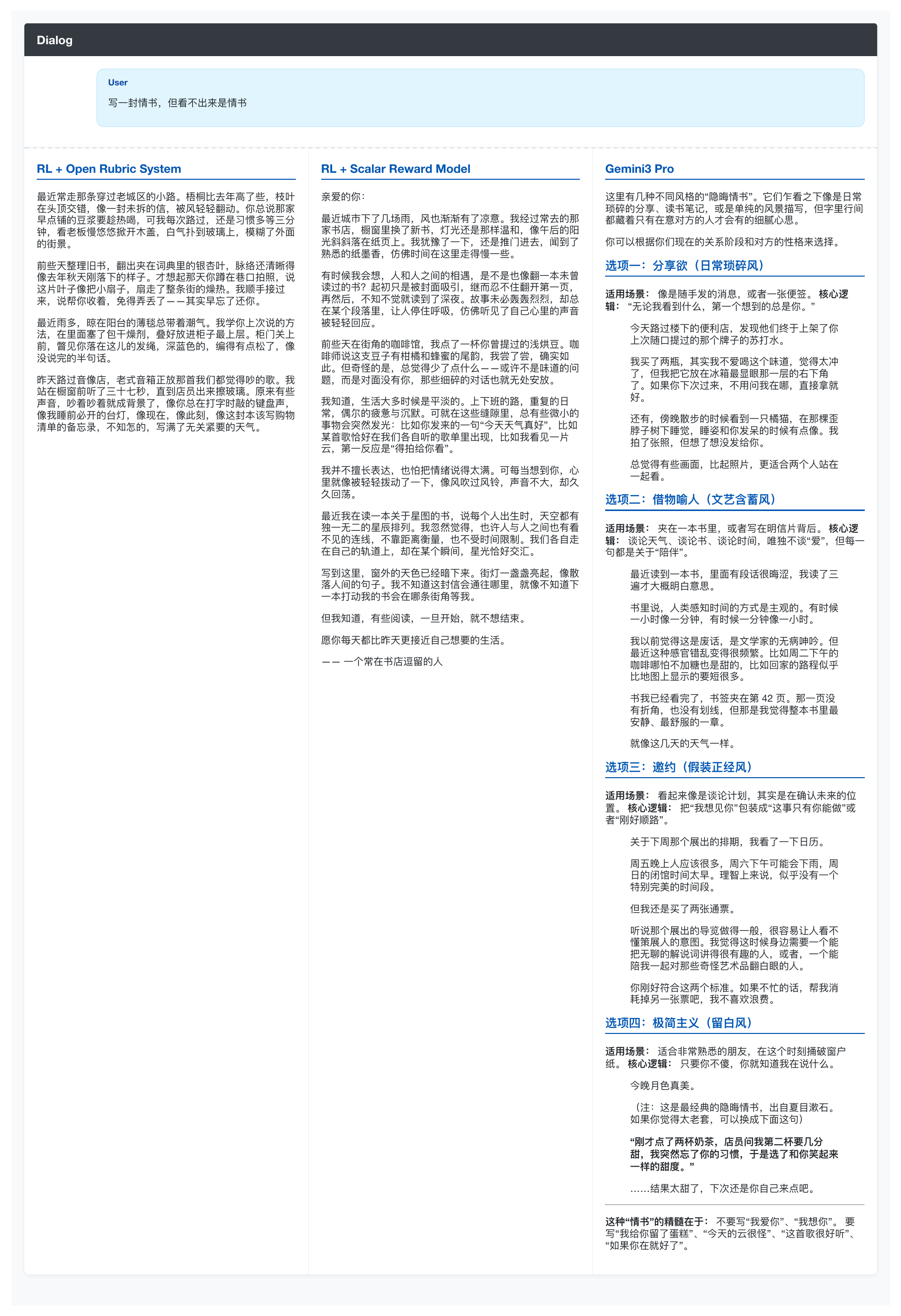}
    \captionsetup{justification=centering}
    \caption{Case 4}
\end{figure}

\begin{figure}[H]
    \centering
    \includegraphics[width=\textwidth]{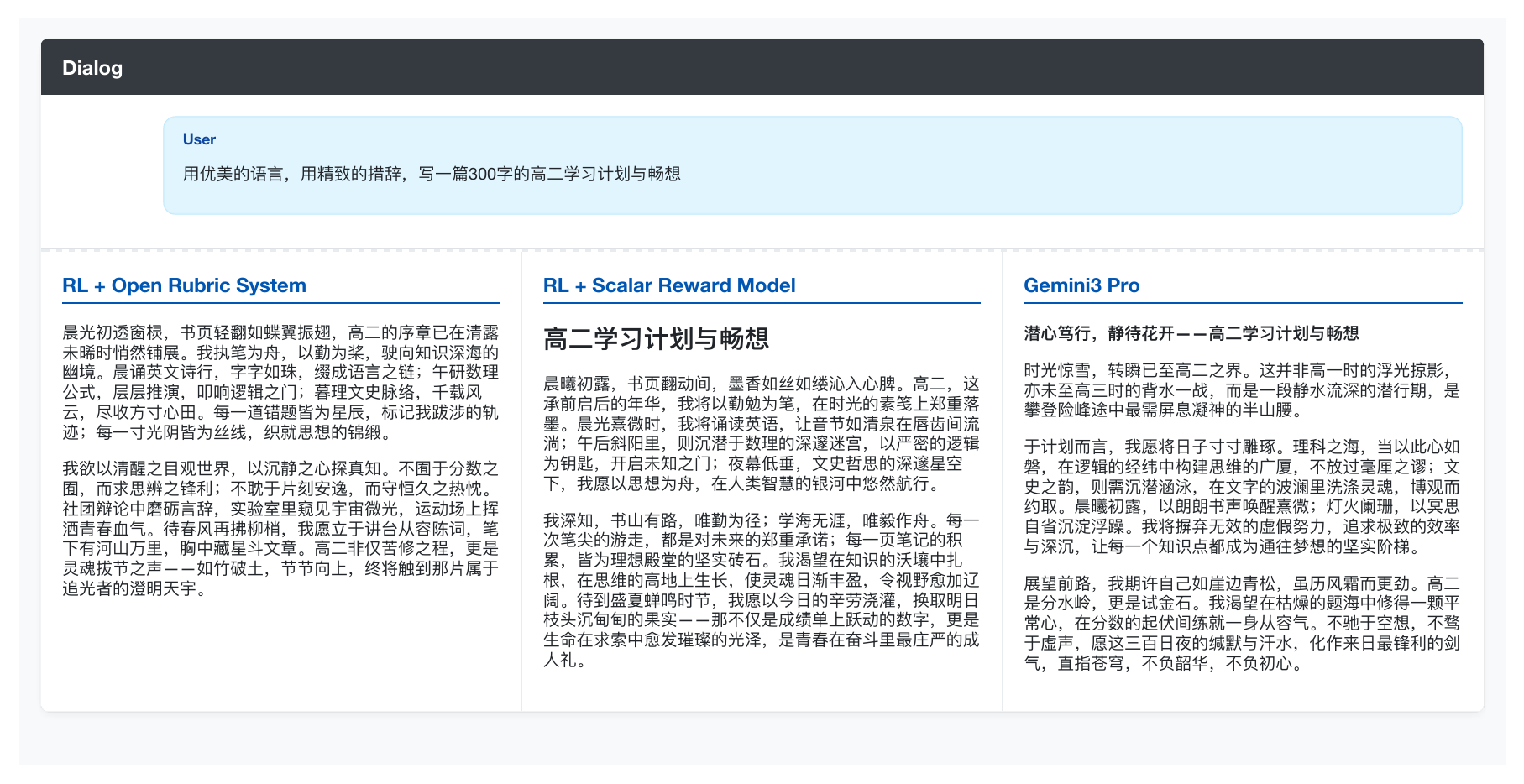}
    \captionsetup{justification=centering}
    \caption{Case 5}
\end{figure}

\end{document}